\documentclass[final]{cvpr}

\usepackage{times}
\usepackage{epsfig}
\usepackage{graphicx}
\usepackage{amsmath}
\usepackage{amssymb}
\usepackage{tabularx}
\usepackage{booktabs}
\usepackage{multirow}
\usepackage{subfigure}
\usepackage{amsmath, bm}
\usepackage{algorithm,algpseudocode}
\usepackage{xcolor,colortbl}

\usepackage[pagebackref=true,breaklinks=true,colorlinks,bookmarks=false]{hyperref}

\def\cvprPaperID{2241} 
\def\confYear{CVPR 2021}

\begin{document}

\title{ST3D: Self-training for Unsupervised Domain Adaptation on 3D Object Detection}

\author{
Jihan Yang$^{1}$\thanks{equal contribution}, \  Shaoshuai Shi$^{2*}$, \ Zhe Wang$^{3,4}$, \  Hongsheng Li$^{2,5}$, \  Xiaojuan Qi$^{1}$\thanks{corresponding author}\\
$^{1}$The University of Hong Kong \ $^{2}$CUHK-SenseTime Joint Laboratory, The Chinese University of Hong Kong \\ $^{3}$SenseTime Research \  $^{4}$Shanghai AI Laboratory \ $^{5}$School of CST, Xidian University \\
{\tt\small \{jhyang, xjqi\}@eee.hku.hk, shaoshuaics@gmail.com, wangzhe@sensetime.com, hsli@ee.cuhk.edu.hk}
}

\maketitle
\pagestyle{empty}
\thispagestyle{empty}

\begin{abstract}
    \vspace{-0.3cm}
    We present a new domain adaptive self-training pipeline, named ST3D, for unsupervised domain adaptation on 3D object detection from point clouds.
    First, we pre-train the 3D detector on the source domain with our proposed random object scaling strategy for mitigating the negative effects of source domain bias. Then, the detector is iteratively improved on the target domain by alternatively conducting two steps, which are the pseudo label updating with the developed quality-aware triplet memory bank and the model training with curriculum data augmentation. These specific designs for 3D object detection enable the detector to be trained with consistent and high-quality pseudo labels and to avoid overfitting to the large number of easy examples in pseudo labeled data. 
    Our ST3D achieves state-of-the-art performance on all evaluated datasets and
    even surpasses fully supervised results on KITTI 3D object detection benchmark. Code will be available at \url{https://github.com/CVMI-Lab/ST3D}.
\end{abstract}

\vspace{-0.5cm}
\section{Introduction}
\vspace{-0.1cm}
3D object detection aims to categorize and localize objects from 3D sensor data ({\eg} LiDAR point clouds) with many applications in autonomous driving, robotics, virtual reality, to name a few. Recently, this field has obtained remarkable advancements~\cite{yan2018second,lang2019pointpillars,shi2019pointrcnn,shi2019points,shi2020pv,shi2021pv} driven by deep neural networks and large-scale human-annotated datasets~\cite{Geiger2012KITTI,sun2020scalability}.

However, 3D detectors developed on one specific domain ({\ie} source domain) might not generalize well to novel testing domains ({\ie} target domains) due to unavoidable domain-shifts arising from different types of 3D sensors, weather conditions and geographical locations, etc. For instance, a 3D detector trained on data collected in USA cities with Waymo LiDAR ({\ie} Waymo dataset~\cite{sun2020scalability}) suffers from a dramatic performance drop (of over $45\%$)~\cite{wang2020train} when evaluated on data from European cities captured by Velodyne LiDAR ({\ie} KITTI dataset~\cite{Geiger2012KITTI}). Though collecting more training data from different domains could alleviate this problem, it unfortunately might be infeasible given various real-world scenarios and enormous costs for 3D annotation. Therefore, approaches to effectively adapting 3D detector trained on labeled source domain to a new unlabeled target domain is highly demanded in practical applications. This task is also known as unsupervised domain adaptation (UDA) for 3D object detection.

In contrast to the intensive studies on UDA of the 2D image setting \cite{ganin2014unsupervised,long2015learning,hoffman2016fcns,chen2018domain,saito2019strong,ge2019mutual,ge2020self}, 
few efforts~\cite{wang2020train} have been made to explore UDA for 3D detection. Meanwhile, the fundamental differences in data structures and network architectures render UDA approaches for image tasks not readily applicable to this problem.  
For DA on 3D detection, while promising results have been obtained in~\cite{wang2020train}, the method requires object size statistics of the target domain, and its efficacy largely depends on  data distributions.
\begin{figure}[t]
    \centering
    \label{fig:performance_comp}
    \includegraphics[width=0.9\linewidth]{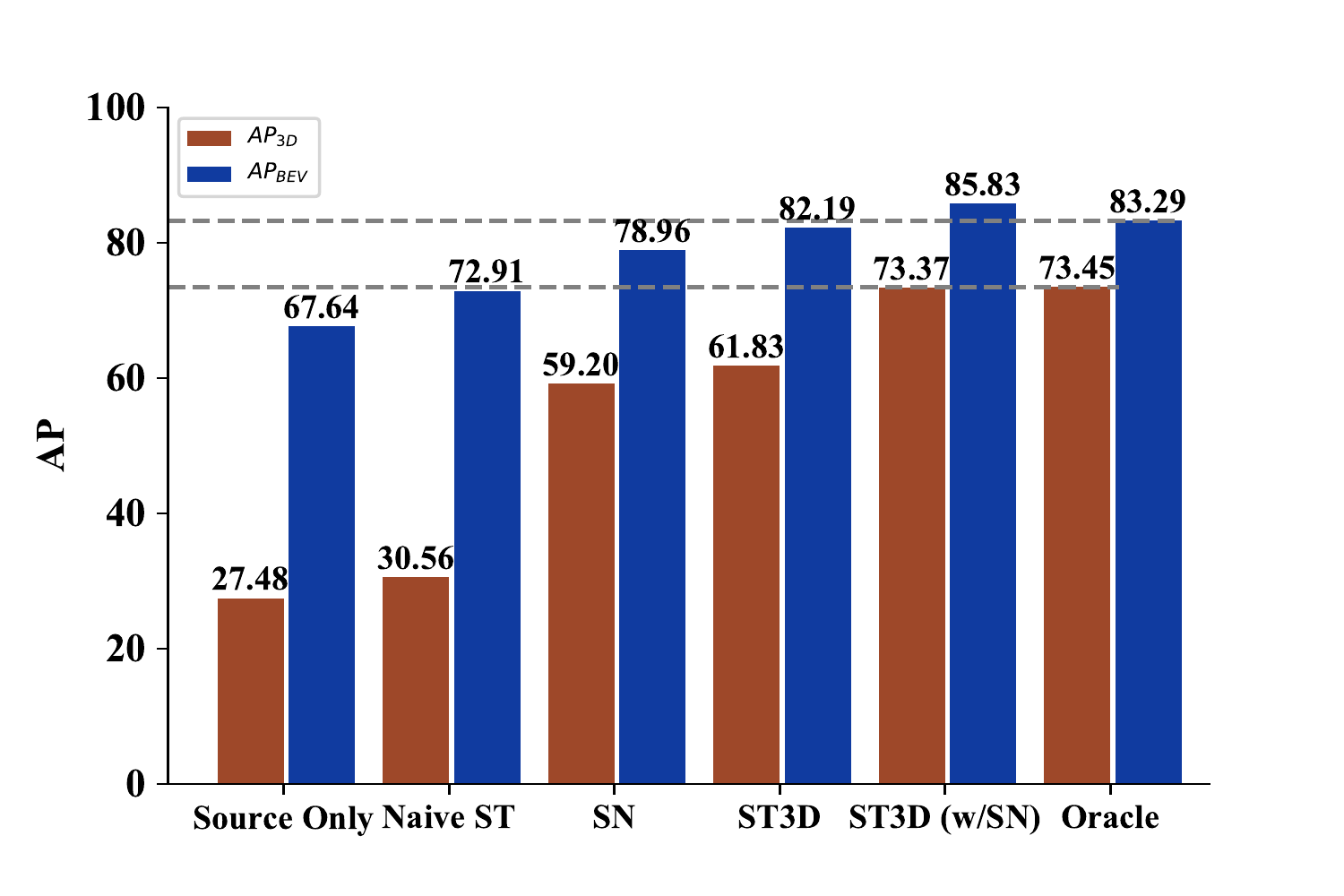}
    \vspace{-0.3cm}
    \caption{Performance of ST3D on Waymo $\rightarrow$ KITTI task using SECOND-IoU \cite{yan2018second}, compared to other unsupervised (\ie source only, naive ST), weakly-supervised (\ie SN \cite{wang2020train}) and fully supervised (\ie oracle) approaches. Dashed line denotes fully supervised target labeled data trained SECOND-IoU.}
    \vspace{-0.55cm}
\end{figure}

Recently, self-training has emerged as a simple and effective technique for UDA, attaining state-of-the-art performance on many image recognition tasks~\cite{zhang2020label,zou2019confidence,khodabandeh2019robust}.
This motivates us to study self-training for UDA on 3D object detection.
Self-training starts from pre-training a model on source labeled data and further iterating between pseudo label generation and model training on unlabeled target data
until convergence is achieved.
The pseudo label for 3D object detection includes oriented 3D bounding boxes for localization and object category information.
Despite of the encouraging results in image tasks, our study illustrates that naive self-training~\cite{xie2020self} does not work well in UDA for 3D detection as shown in Fig.~\ref{fig:performance_comp} (``source only'' {\vs} ``naive ST''). 

In this paper, we propose ST3D, redesigning the self-training pipeline, for UDA on 3D object detection. 
First, in model pre-training, we develop \textit{random object scaling} (ROS), a simple 3D object augmentation technique, randomly scaling the 3D objects to overcome the bias in object size on the labeled source domain. 
Second, for pseudo label generation, we develop a \textit{quality-aware triplet memory bank} (QTMB) which encompasses an IoU-based box scoring criterion to directly assess the quality of pseudo boxes, a triplet box partition scheme to avoid assigning pseudo labels to ambiguous examples, and a memory bank, integrating historical pseudo labels via ensemble and voting, to reduce pseudo label noise and stabilize training.
Finally, in the model training process, we design a \textit{curriculum data augmentation} (CDA) strategy, progressively increasing the intensity of augmentation, to guarantee effective learning at the beginning and gradually simulate hard examples to improve the model, preventing it from overfitting to easy examples -- pseudo-labeled data with high confidence.

Experimental results on four 3D object detection datasets KITTI~\cite{Geiger2012KITTI}, Waymo~\cite{sun2020scalability}, nuSenses~\cite{caesar2020nuscenes}, and Lyft~\cite{lyft2019} demonstrate the effectiveness of our approach, where the performance gaps between source only results  and fully supervised oracle results are closed by a large percentage (16\% $\sim$ 75\% ). Besides, we outperform the existing approach~\cite{wang2020train} by a notable margin on all evaluated settings. It's also noteworthy that our approach even outperforms the oracle results on the Waymo $\rightarrow$ KITTI setting when further combined with existing approach~\cite{wang2020train} as shown in Fig.~\ref{fig:performance_comp}.

\vspace{-0.2cm}
\section{Related Work}
\vspace{-0.2cm}
\noindent
\textbf{3D Object Detection from Point Clouds} aims to localize and classify 3D objects from point clouds, 
which is a challenging task due to the irregularity and sparsity of 3D point clouds.
Some previous work \cite{chen2017multi, ku2018joint,yang2018pixor} directly projects the irregular point clouds to 2D bird-view maps such that the task could be resolved by previous 2D detection methods. 
Another line of research \cite{yan2018second, zhou2018voxelnet,shi2019points,he2020structure, shi2020pv} adopts 3D convolutional networks to learn 3D features from voxelized point clouds, and the extracted 3D feature volumes are also further compressed to bird-view feature maps as the above.  
Recently, point-based approaches \cite{shi2019pointrcnn, yang2019std} propose to directly generate 3D proposals from raw point clouds by adopting PointNet++~\cite{qi2017pointnet++} to extract point-wise features. There are also some other methods \cite{qi2018frustum, wang2019frustum} that utilize 2D images for generating 2D box proposals which are further employed to crop the object-level point clouds for generating 3D bounding boxes. 
In our work, we adopt SECOND~\cite{yan2018second} and PV-RCNN~\cite{shi2020pv} as our 3D object detectors.

\noindent
\textbf{Unsupervised Domain Adaptation}
aims to generalize the model trained on source domain to unlabeled target domains. 
\cite{long2015learning,long2018conditional} explore domain-invariant feature learning by minimizing Maximum Mean Discrepancy \cite{ben2010impossibility}. Inspired by GANs~\cite{goodfellow2014generative}, adversarial learning was employed to align feature distributions across different domains on various 2D vision tasks \cite{ganin2014unsupervised,hoffman2016fcns,chen2018domain,saito2019strong}. Besides, \cite{hoffman2017cycada,zhang2018fully} try to eliminate the domain gap on pixel-level by translating images. Other approaches \cite{saito2017asymmetric,zou2018unsupervised,khodabandeh2019robust,cai2019exploring} utilize the self-training strategy to generate pseudo labels for unlabeled target domains. Saito \etal \cite{saito2018maximum} adopt a two branch classifier to reduce the $\mathcal{H} \Delta \mathcal{H}$ discrepancy. \cite{supancic2013self,choi2019pseudo,chitta2018adaptive} employ curriculum learning~\cite{bengio2009curriculum} and separate cases by their difficulties to realize local sample-level curriculum. Xu \etal \cite{Xu_2019_ICCV} propose a progressive feature-norm  enlarging method to reduce the domain gap. \cite{liu2019transferable,yang2020adversarial} inject feature perturbations to obtain a robust classifier through adversarial training. 

On par with the developments on domain adaptation for image recognition tasks, some recent works also aim to address the domain shift on point clouds for shape classification~\cite{qin2019pointdan} and semantic segmentation~\cite{wu2018squeezeseg,yi2020complete,jaritz2020xmuda}. However, despite of intensive studies on the 3D object detection task~\cite{zhou2018voxelnet, shi2019pointrcnn,yan2018second, shi2019points, yang2019std, shi2020pv}, only very few approaches have been proposed to solve UDA for 3D object detection. Wang {\etal} propose {SN} \cite{wang2020train} to normalize the object size of the source domain leveraging the object statistics of the target domain to close the size-level domain gap. Though the performance has been improved, the method needs the target statistics information, and its effectiveness depends on the source and target data distributions.
In contrast, we propose a novel self-training pipeline for domain adaptive 3D object detection which achieves superior performance on all evaluated settings without target object statistics as a prior. 


\begin{figure*}[t]
	\centering
	\includegraphics[width=\linewidth]{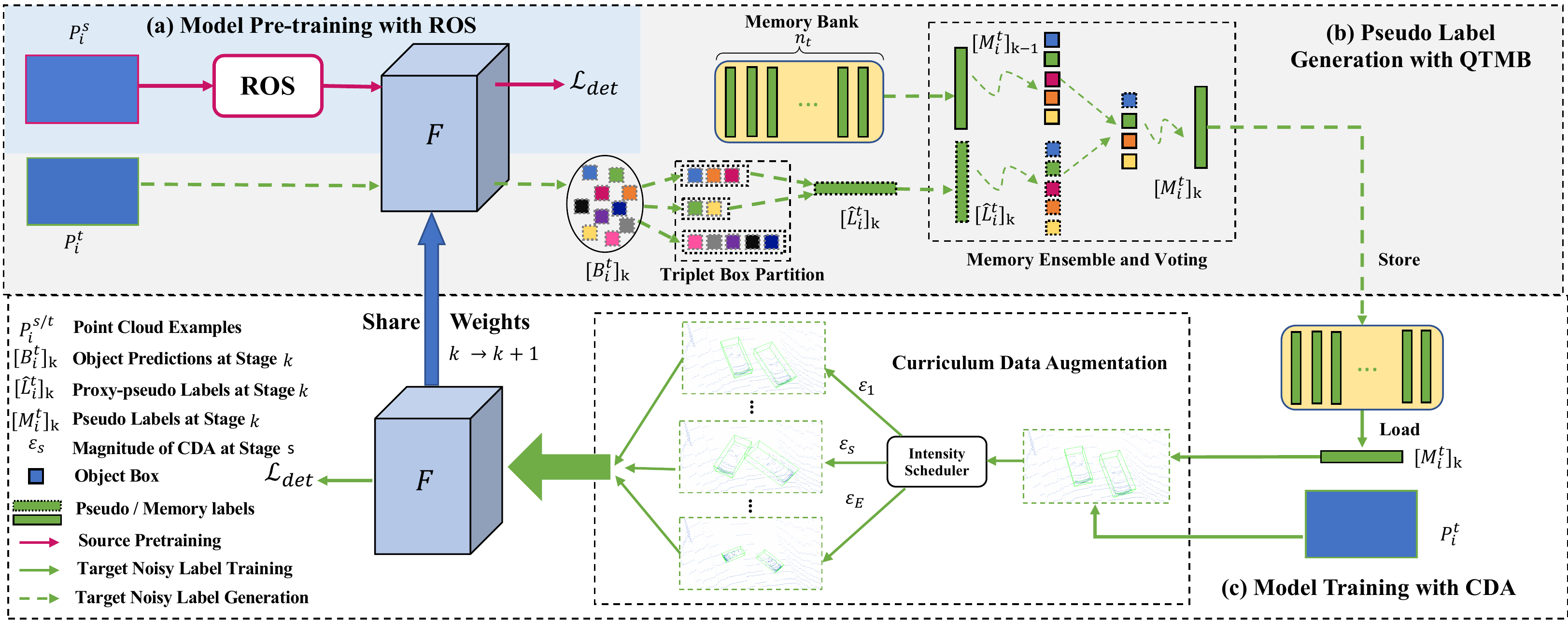}
	\caption{Our ST3D framework consists of three phases: (a) Pre-train the object detector $F$ with ROS in source domain to mitigate object-size bias. (b) Generate high-quality and consistent pseudo labels on target unlabeled data with our QTMB. (c) Train model effectively on pseudo-labeled target data with CDA to progressively simulate hard examples. Best viewed in color.}
    \label{fig:framework}
    \vspace{-0.2cm}
\end{figure*}

\begin{algorithm}[htbp]
	\small
	\caption{Overview of our ST3D. 
	}
	\renewcommand{\algorithmicensure}{ \textbf{Output:}} 
	\label{algo:our_pipeline}
    \begin{algorithmic}[1]
		\Require 
		Source domain labeled data $\{(P^s_i, L^s_i)\}^{n_s}_{i=1}$, and target domain unlabeled data $\{P^t_i\}_{i=1}^{n_t}$.
		\Ensure The object detection model for target domain.
		
		\State {Pre-train} the object detector on $\{(P^s_i, L^s_i)\}^{n_s}_{i=1}$ with ROS as detailed in Sec.~\ref{sec:pretrain}. \label{algo:line_ros}
		\State Utilize the current model to generate {raw object proposals} $[B^{t}_i]_k$ for every sample $P_i^t$, where $k$ is the current number of times for pseudo label generation.
		\State Generate quality-aware pseudo labels $[\hat{L}^{t}_i]_k$ by triplet box partition given $[B^{t}_i]_k$ in Sec.~\ref{sec:triplet}. \label{algo:line_triplet}
		\State Update the memory ({\ie} pseudo labels) $[M_i^{t}]_k$ given pseudo labels $[\hat{L}^{t}_i]_k$ from the detection model and historical pseudo labels $[M_i^{t}]_{k-1}$ $([M_i^{t}]_0 = \emptyset)$ in the memory with memory ensemble-and-voting (MEV) as elaborated in Sec.~\ref{sec:memoryensemble}.
		The memory bank $\{[M^{t}_i]_k\}_{i=1}^{n_t}$ contains the pseudo labels for all unlabeled examples.
		\State Train the model on $\{P_i^t, [M_i^{t}]_k\}_{i=1}^{n_t} $ with CDA for several epochs as detailed in Sec.~\ref{sec:self_training}. \label{algo:line_CDA}
		\State Go back to Line 2 until convergence.
	\end{algorithmic}
\end{algorithm}

\section{Method}
\vspace{-0.2cm}
\subsection{Overview}
\vspace{-0.1cm}
Our goal is to adapt a 3D object detector trained on source labeled data $\{(P^s_i, L^s_i)\}^{n_s}_{i=1}$ of $n_s$ samples to unlabeled target domain given target unlabeled data $\{P^t_i\}_{i=1}^{n_t}$ of $n_t$ samples. Here, $P^s_i$ and $L^s_i$ represent the $i$-th source input point cloud and its corresponding label.  
$L^s_i$ contains the category and 3D bounding box information for each object in the $i$-th point clouds, and each box is parameterized by its size $(l, w, h)$, center $(c_x, c_y, c_z)$, and heading angle $\theta$. Similarly, $P^t_i$ denotes the $i$-th unlabeled target point cloud.

In this section, we present ST3D, a self-training framework for adapting the 3D detector trained on source domain to target domain, which is shown in Fig.~\ref{fig:framework} and described in Algo.~\ref{algo:our_pipeline}. Starting from pre-training a detector on source labeled data with random object scaling (ROS) (see Fig.~\ref{fig:framework} (a)), ST3D alternates between generating pseudo labels for target data via quality-aware triplet memory bank (QTMB) (see Fig.~\ref{fig:framework} (b)) and training the detector with our curriculum data augmentation (CDA) (see Fig.~\ref{fig:framework} (c)) until convergence.

\subsection{Model Pre-training with ROS}\label{sec:pretrain}
\vspace{-0.1cm}
Our ST3D starts from training a 3D object detector on labeled source data  $\{(P^s_i, L^s_i)\}^{n_s}_{i=1}$. The pre-trained model learns how to perform 3D detection on source labeled data and is further adopted to initialize object predictions for the target domain unlabeled data. 

\noindent
\textbf{Motivation.}~
However, despite of the useful knowledge, the pre-trained detector also learns the bias from the source data, such as object size and point densities due to domain shift.
Among them, the bias in object size has direct negative impacts on 3D object detection, and results in incorrect size for pseudo-labeled target domain bounding boxes. This is also in line with the findings in~\cite{wang2020train}. To mitigate the issue, we propose a very simple yet effective per-object augmentation strategy, {\ie} \textit{random object scaling} (ROS), fully leveraging the high degree of freedom of 3D spaces.

\noindent
\textbf{Random Object Scaling.}~
Given an annotated 3D bounding box with size $(l, w, h)$, center $(c_x, c_y, c_z)$ and heading angle $\theta$, ROS scales the box in the length, width and height dimensions with random scale factors $(r_l, r_w, r_h)$ through transforming the points inside the box. 
We denote the points inside the box as $\{p_i\}_{i=1}^{n_p}$ with a total of $n_p$ points, and the coordinate of $p_i$ is represented as $(p_i^x, p_i^y, p_i^z)$.
First, we transform the points to the local coordinate system of the box along its length, width and height dimensions via
\vspace{-0.15cm}
\begin{equation}
\vspace{-0.2cm}
\begin{small}
\begin{aligned}
(p_i^l, p_i^w, p_i^h) &= (p_i^x - c_x, p_i^y - c_y, p_i^z - c_z) \cdot R, \\
 R &=
	\begin{bmatrix}
	\cos \theta & -\sin \theta & 0\\
	\sin \theta & \cos \theta & 0\\
	0 & 0 & 1\\
	\end{bmatrix},
    \end{aligned}
\end{small}
\end{equation}
where $\cdot$ is matrix multiplication.
Second, to derive the scaled object, the point coordinates inside the box are scaled to be $(r_lp_i^l, r_wp_i^w, r_hp_i^h)$ with object size $(r_ll, r_ww, r_hh)$.
Third, to derive the augmented data $\{p_i^{\text{aug}}\}_{i=1}^{n_p}$, the points inside the scaled box are transformed back to the ego-car coordinate system and shifted to the center $(c_x, c_y, c_z)$ as
\vspace{-0.25cm}
\begin{equation}
\vspace{-0.1cm}
p_i^\text{aug} = (r_lp_i^l, r_wp_i^w, r_hp_i^h)\cdot R^T + (c_x, c_y, c_z).
\end{equation}
Albeit simple, ROS effectively simulates objects with diverse object sizes to address the size bias and hence facilitates to train size-robust detectors that produce more accurate initial pseudo boxes for subsequent self-training.

\subsection{Pseudo label Generation with QTMB}\label{sec:memory_bank}

With the trained detector, the next step is to generate pseudo labels for the unlabeled target data.
Given the target sample $P_i^t$, the output $B_i^t$ of the object detector is a group of predicted boxes containing category confidence scores, regressed box sizes, box centers and heading angles, where non-maximum-suppression (NMS) has already been conducted to remove the redundant boxes. 
For clarity, we call $B_i^t$ as the object predictions for a scene.

\begin{figure}[t]
    \centering
    \includegraphics[width=\linewidth]{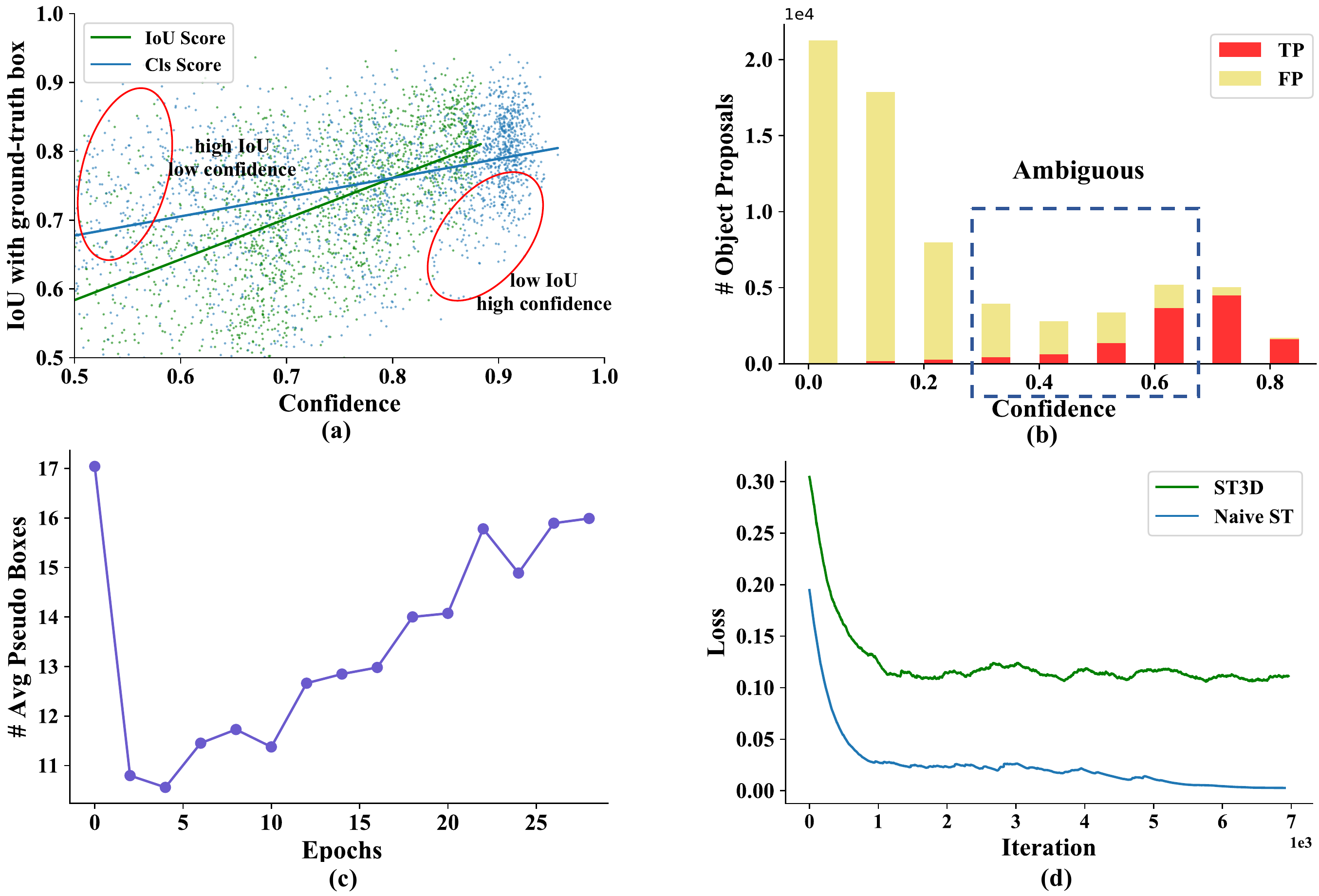}
    \caption{(a) Correlation between confidence value and box IoU with ground-truth (b) Lots of boxes with medium confidence may be assigned with ambiguous labels.
    (c) The average number of pseudo boxes fluctuates at different epochs. (d) Training loss curve comparison between naive ST and our ST3D with CDA.}
    \label{fig:motivations}
    \vspace{-0.3cm}
\end{figure}

\noindent
\textbf{Motivation.}
Different from classification and segmentation tasks, 3D object detection needs to jointly consider the classification and localization information, which poses great challenges for high-quality pseudo label generation. 
First, the confidence of object category prediction may not necessarily reflect the precision of location as shown by the blue line in Fig.~\ref{fig:motivations} (a). 
Second, the fraction of false labels is much increased in confidence score intervals with medium values as illustrated in Fig.~\ref{fig:motivations} (b). 
Third, model fluctuations induce
inconsistent pseudo labels as demonstrated in Fig.~\ref{fig:motivations}~(c). The above factors will undoubtedly have negative impacts on the pseudo-labeled objects, leading to noisy supervisory information and instability for self-training. 

To address the above challenges, we design \textit{quality-aware triplet memory bank} (QTMB) to parse object predictions to pseudo labels for self-training.
The memory bank at the $k$-th pseudo label generation stage, denoted as $\{[M_i^t]_k\}_{i=1}^{n_t}$, contains  pseudo labels for all target domain data.
$\{[M_i^t]_k\}_{i=1}^{n_t}$ is derived by combining pseudo labels $\{[\hat{L}_i^t]_k\}_{i=1}^{n_t}$ from the object detector and historical pseudo labels $\{[M_i^t]_{k-1}\}_{i=1}^{n_t}$ in the memory via ensemble and voting. 
Meanwhile, given the object predictions $\{B_i^t\}_{i=1}^{t}$ from the detector, $\{[\hat{L}_i^t]_k\}_{i=1}^{n_t}$ is constructed with an IoU-based scoring criterion to ensure the localization quality and a triplet box partition scheme to safely avoid assigning different labels to objects predictions with ambiguous confidence.   
To differentiate pseudo labels $\{[\hat{L}_i^t]_k\}_{i=1}^{n_t}$ from the object detector and pseudo labels $\{[M_i^t]_{k-1}\}_{i=1}^{n_t}$  in the memory,  we call $\{[\hat{L}_i^t]_k\}_{i=1}^{n_t}$ ``proxy-pseudo label'' in what follows.
\
\subsubsection{Proxy-pseudo Labels from the Object Detector} \label{sec:triplet}
Firstly, to obtain high-quality and accurate proxy-pseudo labels $\{[\hat{L}_i^t]_k\}_{i=1}^{n_t}$ from the  detection model, we introduce an IoU-based quality-aware criterion to directly assess the quality of the box, and a triplet box partition scheme to reduce noise from ambiguous objects predictions.

%
\noindent
\textbf{IoU-based Quality-aware Criterion for Scoring.}
To assess the localization quality of pseudo labels, we propose to augment the original object detection model with a lightweight IoU regression head. 
Specifically, given the feature derived from RoI pooling, we append two fully connected layers to directly predict the 3D box IoU between RoIs and their ground truths (GTs) or pseudo labels. 
A sigmoid function is adopted to map the output into range $[0,1]$.
During model training, the IoU branch is optimized by a binary cross entropy loss as 
\vspace{-0.2cm} 
\begin{equation}
    \vspace{-0.15cm}
	\mathcal{L}_{iou}=-\hat{u}\log u - (1 - \hat{u})\log (1-u),
\end{equation}
where $u$ is the predicted IoU and $\hat{u}$ is the IoU between the ground truth (or pseudo label) box and the predicted 3D box.
The correlation between the IoU score and localization quality (see green line in Fig.~\ref{fig:motivations} (a)) is much increased in comparison with the classification confidence.
Though IoU regression has been tried to improve supervised image object detection performance~\cite{huang2019mask,bolya2019yolact++}, to the best of our knowledge, we are the first to demonstrate that it can serve as a good criterion to assess the quality of pseudo box for UDA self-training with encouraging results.

\noindent
\textbf{Triplet Box Partition to Avoid Ambiguous Samples.}
Now, we are equipped with a better IoU-based quality assessment criterion and object predictions $[B^t_i]_k$ (for the $i$-th sample at stage $k$) from the detector after NMS.
Here, we present a triplet box partition scheme to obtain the proxy-pseudo labels $[\hat{L}_i^t]_k$  to avoid assigning labels to ambiguous examples. Given an object box $b$ from $[B^t_i]_k$ with IoU prediction score $u_b$, we create a  margin $[T_{\text{neg}}, T_\text{pos}]$ to ignore boxes with score $u_b$ inside the margin, preventing them from contributing to training, as follows:
\vspace{-0.2cm}
\begin{align}
\text{state}_b\! =\! \!\left\{
\begin{tabular}{@{}l@{}}
$\text{Positive} \text{ (Store to } [\hat{L}_i^t]_k \text{)}, \  {T_{\text{pos}} \leq u_b}, $\\
$\text{Ignored} \text{ (Store to } [\hat{L}_i^t]_k \text{)} , \  {T_{\text{neg}} \leq u_b < T_{\text{pos}}},$\\
$\text{Negative} \text{ (Discard)} , \ \ \ \ \ \ \ \ \ \ \ \ \  {u_b < T_{\text{neg}}}.$
\end{tabular}
\right. 
\end{align}
If $\text{state}_b$ is positive, $b$ will be cached into $[\hat{L}_i^t]_k$ as a positive sample with its category label and pseudo box. Similarly, the ignored boxes will also be incorporated into the $[\hat{L}_i^t]_k$ to identify regions that should be ignored during model training due to its high uncertainty. Box $b$ with negative $\text{state}_b$ will be discarded, corresponding to backgrounds.

Our triplet box partition scheme reduces noisy pseudo labels from ambiguous boxes and ensures the quality of pseudo-labeled boxes. To be noted, objects on the ignored regions may be evoked later if their scores are improved. 
\vspace{-0.2cm}
\subsubsection{Memory Update and Pseudo Label Generation}~\label{sec:memoryensemble}
Here, we combine proxy-pseudo labels $\{[\hat{L}_i^t]_k\}_{i=1}^{n_t}$ at stage $k$ and the historical pseudo  labels $\{[M_i^{t}]_{k-1}\}_{i=1}^{n_t}$ $([M_i^{t}]_0 = \emptyset)$ in the memory bank via memory ensemble and voting. The outputs are the updated pseudo labels $\{[M_i^{t}]_{k}\}_{i=1}^{n_t}$ that also serve as the labels for the subsequent model training.
During this memory update process, each pseudo box $b$ from $[\hat{L}_i^t]_k$ and $[M_i^{t}]_{k-1}$ has three attributes $(u_b, \text{state}_b, \text{cnt}_b)$, which are the confidence score, state (positive or ignored)  and an unmatched memory counter (UMC) (for memory voting), respectively.
We assume that $[\hat{L}_i^t]_k$ contains $n_l$ boxes denoted as $[\hat{L}_i^t]_k = \{(u_l, \text{state}_l, \text{cnt}_l)_j^k\}_{j=1}^{n_l}$ and  $[M_i^t]_{k-1}$ has $n_m$ boxes represented as $[M_i^t]_{k-1} = \{(u_m, \text{state}_m, \text{cnt}_m)_j^{k-1}\}_{j=1}^{n_m}$.


\begin{figure}[t]
    \centering
    \includegraphics[width=0.9\linewidth]{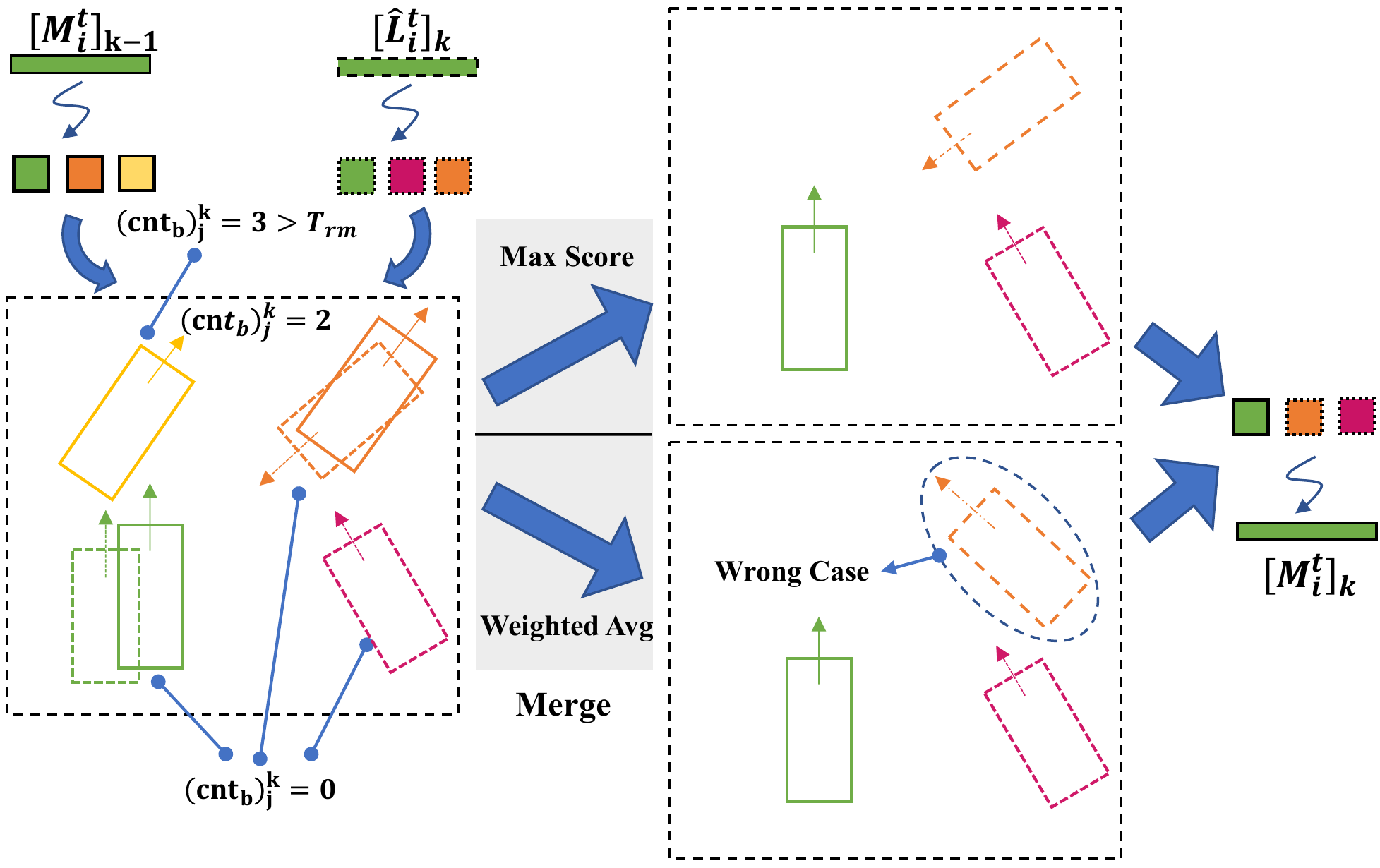}
    \caption{An instance of memory ensemble and voting (MEV). Given proxy-pseudo labels and historical memory labels, MEV automatically matches and merges boxes while ignoring or discarding successively unmatched boxes. The weighted average boxes merging strategy could produce wrong final box for boxes with very different heading angles.}
    \label{fig:memory_ensemble}
    \vspace{-0.2cm}
\end{figure}
 
\noindent
\textbf{Memory Ensemble.}
Instead of directly replacing $[M_i^t]_{k-1}$ with the latest proxy-pseudo labels $[\hat{L}_i^t]_k$, we propose the memory ensemble operation to combine $[M_i^t]_{k-1}$ and $[\hat{L}_i^t]_k$ to produce more consistent and high-quality pseudo labels.

The memory ensemble operation matches two object boxes with similar locations, sizes and angles from $[M_i^t]_{k-1}$ and $[\hat{L}_i^t]_k$, and merges them to produce a new object box. 
By default, we adopt the consistency ensemble strategy for box matching. 
Specifically, it calculates the pair-wise 3D IoU matrix $A = \{a_{jv}\} \in \mathbb{R}^{n_m \times n_l}$ between each box in $[M_i^t]_{k-1}$ and each box in $[\hat{L}_i^t]_k$. 
For the $j$-th object box in $[M_i^t]_{k-1}$, its matched box index $\hat{j}$ in $[\hat{L}_i^t]_k$ is derived by,
\vspace{-0.1cm}
\begin{equation}
    \vspace{-0.2cm}
    \hat{j} = \text{argmax}_j~ (a_{jv}), ~{v=1, \cdots, n_l}.
\end{equation}
Note that if $a_{j\hat{j}} < 0.1$, we denote each of these two paired boxes as unmatched boxes that will be further processed by the memory voting operation.

We assume the successfully matched pair-wise object boxes as  $(u_l, \text{state}_l, \text{cnt}_l)_{\hat{j}}^{k}$ and 
$(u_m, \text{state}_m, \text{cnt}_m)_{j}^{k-1}$. 
They are further merged to cache the pseudo labeled box with a higher confidence value into the $[M_i^{t}]_{k}$ and update its corresponding attributes as
\vspace{-0.1cm}
\begin{equation}
\vspace{-0.1cm}
\label{eq:me_state_update}
\begin{small}
(u_m, \text{state}_m, 0)_j^{k} = \!\left\{\!
\begin{array}{l}\!
\!(u_l, \text{state}_l, \text{cnt}_l)_{\hat{j}}^{k} , \ \ \ \ \text{if} ~{u_m \!\leq\! u_l}, \\\!
\!(u_m, \text{state}_m, \text{cnt}_m)_j^{k-1}, \text{otherwise}, 
\end{array}
\right.
\end{small}
\end{equation}
%
Here, we adopt to choose box instead of a weighted combination is because weighted combination has the potential to produce an unreasonable final box if the matched boxes have very different heading angles (see Fig.~\ref{fig:memory_ensemble} ``wrong case'').
We also explore two alternative strategies for box matching, which are discussed in Sec.~\ref{abl:mev}.

\noindent
\textbf{Memory Voting.}
The memory ensemble operation can effectively select better matched pseudo boxes. 
However, it cannot handle the unmatched pseudo boxes from either $[M_i^t]_{k-1}$ or $[\hat{L}_i^t]_k$.
As the unmatched boxes often contain both false positive boxes and high-quality true positive boxes, either caching them into the memory or discarding them all is suboptimal.  
To address the above problem, we propose a novel memory voting approach, which leverages history information of unmatched object boxes to robustly determine their status (\textbf{cache}, \textbf{discard} or \textbf{ignore}).
%
%
 For the $j$-th {unmatched} pseudo boxes $b$ from $[M_i^t]_{k-1}$ or $[\hat{L}_i^t]_k$, its UMC $(\text{cnt}_b)_j^{k}$ will be updated as follows: 
\vspace{-0.3cm}
\begin{equation}
\vspace{-0.2cm}
\begin{small}
(\text{cnt}_b)^k_{j} = \left\{
\begin{array}{lll}
0 &, & \text{if } b \in [\hat{L}_i^t]_k,  \\
(\text{cnt}_b)_j^{k-1} + 1 & , & \text{if }  b \in [M_i^t]_{k-1},\\
\end{array}
\right.
\end{small}
\end{equation}
We update the UMC for unmatched boxes in $[M_i^t]_{k-1}$ by adding $1$ and initialize  the UMC of the newly generated boxes in  $[\hat{L}_i^t]_k$ as 0.
The UMC records the successive unmatched times of a box, which are combined with two thresholds $T_{\text{ign}}$ and $T_{\text{rm}}$ 
($T_{\text{ign}}=2$ and $T_{\text{rm}}=3$ by default) 
to select the subsequent operation for unmatched boxes as
\vspace{-0.2cm}
\begin{equation}
\vspace{-0.1cm}
\begin{small}
\left\{\!
\begin{array}{llc}\!
\text{Discard} &, & (\text{cnt}_b)^k_{j} \geq T_{\text{rm}} ,  \\\!
\text{Ignore}\ (\text{Store to} [M_i^{t}]_{k}) & , & T_{\text{ign}} \leq (\text{cnt}_b)^k_{j} < T_{\text{rm}} , \\
\!\text{Cache}\ (\text{Store to} [M_i^{t}]_{k})& , & (\text{cnt}_b)^k_{j} < T_{\text{ign}} .\\
\end{array}
\right.
\end{small}
\end{equation}
Benefited from our memory voting, we could generate more robust and consistent pseudo boxes by caching the occasionally unmatched box in the memory bank.


\subsection{Model training with CDA}\label{sec:self_training}
Our proposed QTMB can produce consistent and stable pseudo labels $[M_i^{t}]_k$ for the $i$-th point clouds. 
Now, the detection model can be trained on $\{P_i^t, [M_i^{t}]_k\}_{i=1}^{n_t} $ at stage $k$ as described in Algo.~\ref{algo:our_pipeline} (Line~\ref{algo:line_CDA}).

\noindent
\textbf{Motivation.} However, our observations show that most of positive pseudo boxes are easy examples since they are generated from previous high-confident object predictions. 
Consequently, during training, model is prone to overfitting to these easy examples with low loss values (see Fig.~\ref{fig:motivations}~(d)), unable to further mine hard examples to improve the detector~\cite{bengio2009curriculum}.
To prevent model from being trapped by bad local minimal, strong data augmentations could be an alternative to generate diverse and potentially hard examples to improve the model.
However, this might confuse the learner and hence be harmful to model training at the initial stage.

\noindent
\textbf{Curriculum Data Augmentation.} 
Motivated by the above observation, we design a curriculum data augmentation (CDA) strategy to progressively increase the intensity $\epsilon$ of data augmentation and gradually generate increasingly harder examples to facilitate improving the model and ensure effective learning at the early stages. 

To progressively increase the intensity $\epsilon$ of data augmentations $\{D_i\}^{n_d}_{i=1}$ with $n_d$ types (\ie world coordinate system transformation and per-object coordinate system transformation), we design a  multi-step intensity scheduler with initial intensity $\epsilon_0^i$ for the $i$-th data augmentation.
Specifically, we split the total training epochs into $E$ stages.
After each stage, the data augmentation intensity is multiplied by an enlarging ratio $\alpha$ ($\alpha > 1$, we use $\alpha=1.2$ by default).
Thus, the data augmentation intensity for $i$-th data augmentation at stage $s$ ($1 \leq s \leq E$) is derived as $\epsilon_s^i = \epsilon_0^i {\alpha}^{s-1}$.
Hence, the random sampling range of the $i$-th data augmentation could be calculated as follows:
\begin{equation}
\vspace{-0.2cm}
\begin{small}
\left\{
\begin{array}{llc}
[-\epsilon_s^i,  \epsilon_s^i]  &, & \  \text{if} \ D_i \text{ belongs to rotation},  \\

[1-\epsilon_s^i,  1+\epsilon_s^i] &, &\text{if} \ D_i \text{ belongs to scaling}.\\
\end{array}
\right.
\end{small}
\end{equation}

\noindent CDA enables the model to learn from the challenging samples while making the difficulty of examples be within the capability of the learner during the whole training process.
\section{Experiments}

\subsection{Experimental Setup}

\begin{table*}[htbp]
    \centering
    \begin{small}
        \begin{tabular}{c|c|c|c|c|c}
            \bottomrule[1pt]
            \multirow{2}{*}{Task} & \multirow{2}{*}{Method}  & \multicolumn{2}{c|}{SECOND-IoU} & \multicolumn{2}{c}{PV-RCNN} \\
            \cline{3-6} 
            &  & $\text{AP}_{\text{BEV}}$ / $\text{AP}_{\text{3D}}$ & Closed Gap & $\text{AP}_{\text{BEV}}$ / $\text{AP}_{\text{3D}}$ & Closed Gap \\
            \hline
            \multirow{5}{*}{Waymo $\rightarrow$ KITTI} & Source Only  & 67.64 / 27.48 & - & 61.18 / 22.01 & - \\
            & SN \cite{wang2020train} & 78.96 / 59.20 & +72.33\% / +69.00\% & 79.78 / 63.60 & +66.91\% / +68.76\% \\
            \cline{2-6}
            & ST3D & 82.19 / 61.83 & +92.97\% / +74.72\% & 84.10 / 64.78 & +82.45\% / +70.71\% \\
            & ST3D (w/ SN) & \textbf{85.83} / \textbf{73.37} & +116.23\% / +99.83\% & \textbf{86.65} / \textbf{76.86} & +91.62\% / +90.68\% \\
            \cline{2-6}
            & Oracle &  83.29 / 73.45 & - & 88.98 / 82.50 & - \\
            \toprule[1pt]
            \bottomrule[1pt]
            \multirow{5}{*}{Waymo $\rightarrow$ Lyft} & Source Only  & 72.92 / 54.34 & - &  75.49 / 58.53 & - \\
            & SN \cite{wang2020train}  & 72.33 / 54.34 & {\large-}05.11\% / +00.00\% & 72.82 / 56.64 & {\large-}24.34\% / {\large-}14.36\% \\
            \cline{2-6}
            & ST3D  & 76.32 / \textbf{59.24} & +29.44\% / +33.93\% &  \textbf{77.68} / \textbf{60.53} & +19.96\% / +15.20\% \\
            & ST3D (w/ SN) & \textbf{76.35} / 57.99 & +15.71\% / +17.81\% & 74.95 / 58.54 & {\large-}04.92\% / +00.08\% \\
            \cline{2-6}
            & Oracle &  84.47 / 68.78 & - &  86.46 / 71.69 & - \\
            \toprule[1pt]
            \bottomrule[1pt]
            \multirow{5}{*}{Waymo $\rightarrow$ nuScenes} & Source Only  & 32.91 / 17.24 & - & 34.50 / 21.47 & - \\
            & SN \cite{wang2020train} & 33.23 / 18.57 & +01.69\% / +07.54\% &  34.22 / 22.29 & {\large-}01.50\% / +04.80\% \\
            \cline{2-6}
            & ST3D & \textbf{35.92} / 20.19 & +15.87\% / +16.73\% & 36.42 / 22.99 & +10.32\% / +08.89\% \\
            & ST3D (w/ SN) & 35.89 / \textbf{20.38}  & +15.71\% / +17.81\% & \textbf{36.62} / \textbf{23.67} & +11.39\% / +12.87\% \\
            \cline{2-6}
            & Oracle &  51.88 / 34.87 & - &  53.11 / 38.56 & - \\
            \toprule[1pt]
            \bottomrule[1pt]
            \multirow{5}{*}{nuScenes $\rightarrow$ KITTI} & Source Only & 51.84 / 17.92  & - & 68.15 / 37.17 & - \\
            & SN \cite{wang2020train} & 40.03 / 21.23 & {\large-}37.55\% / +05.96\%  & 60.48 / 49.47 & {\large-}36.82\% / +27.13\% \\
            \cline{2-6}
            & ST3D  & 75.94 / 54.13 & +76.63\% / +59.50\% & 78.36 / 70.85 & +49.02\% / +74.30\% \\
            & ST3D (w/ SN) &  \textbf{79.02} / \textbf{62.55} & +86.42\% / +80.37\% & \textbf{84.29} / \textbf{72.94} & +77.48\% / +78.91\% \\
            \cline{2-6}
            & Oracle & 83.29 / 73.45 & - & 88.98 / 82.50 & - \\
            \toprule[0.8pt]
        \end{tabular}
    \end{small}
    \caption{Result of different adaptation tasks. We report $\text{AP}_{\text{BEV}}$ and $\text{AP}_{\text{3D}}$ of the car category at IoU = 0.7 as well as the domain gap closed by various approaches along Source Only and Oracle. The reported AP is moderate case for the adaptation tasks for to KITTI tasks, and is the overall result for other adaptation tasks. We indicate the best adaptation result by \textbf{bold}.}
    \vspace{-0.3cm}
    \label{tab:SOTAcomparison}
\end{table*}

\noindent
\textbf{Datasets.}~
We conduct experiments on four widely used autonomous driving datasets: KITTI~\cite{Geiger2012KITTI}, Waymo~\cite{sun2020scalability}, nuSenses~\cite{caesar2020nuscenes}, and Lyft~\cite{lyft2019}. Our experiments lie in two aspects: Adaptation from label rich domains to label insufficient domains (i.e., Waymo to other datasets) and across domains with different number of the LiDAR beams (i.e., Waymo $\rightarrow$ nuScenes and nuScenes $\rightarrow$ KITTI). 

\noindent
\textbf{Comparison Methods.}~
We compare ST3D with three methods: $(i)$ \textbf{Source Only} indicates directly evaluating the source domain pre-trained model on the target domain. $(ii)$ \textbf{SN} \cite{wang2020train} is the SOTA domain adaptation method on 3D object detection with target domain statistical object size as extra information. $(iii)$ \textbf{Oracle} indicates the fully supervised model trained on the target domain.

\noindent
\textbf{Evaluation Metric.}~
We follow \cite{wang2020train} and adopt the KITTI evaluation metric for evaluating our methods on the commonly used car category (also named vehicle in the Waymo Open Dataset). 
We evaluate all settings on ring view point clouds since it is more useful in real-world applications, except for the KITTI dataset which only provides the annotations in the front view.
We follow the official KITTI evaluation metric and report the average precision (AP) over 40 recall positions, and the IoU thresholds are 0.7 for both the bird's eye view (BEV) IoUs and 3D IoUs. To further demonstrate the effectiveness of different methods for adaptation, we also report how much the performance gap between Source Only to Oracle is closed, which is represented as \textbf{closed gap}
$=\frac{\text{AP}_{\text{model}} - \text{AP}_{\text{source only}}}{\text{AP}_{\text{oracle}} - \text{AP}_{\text{source only}}} \times 100\%.$

\noindent 
\textbf{Implementation Details.}~
We validate our proposed ST3D on two detection backbones SECOND \cite{yan2018second} and PV-RCNN~\cite{shi2020pv}. Specifically, we improve the SECOND detector with an extra IoU head to estimate the IoU between the object proposals and their GTs, and name this detector as SECOND-IoU. We adopt the training settings of the popular point cloud detection codebase OpenPCDet \cite{openpcdet2020} to pre-train our detectors on the source domain with our proposed random object scaling (ROS) data augmentation strategy. 
For the following target domain self-training stage, we use Adam \cite{kingma2014adam} with learning rate $1.5 \times 10^{-3}$ and one cycle scheduler to finetune the detectors for 30 epochs with curriculum data augmentation (CDA). We update the pseudo label with QTMB after every 2 epochs.
For all the above datasets, the detection range is set to $[-75.2, 75.2]m$ for $X$ and $Y$ axes, and $[-2, 4]m$ for $Z$ axis (the origins of coordinates of different datasets have been shifted to the ground plane). We set the voxel size of both SECOND-IoU and PV-RCNN to $(0.1m, 0.1m, 0.15m)$ on all datasets.

During both the pre-training and self-training processes, we adopt the widely adopted data augmentation, including random flipping, random world scaling, random world rotation, random object scaling and random object rotation. CDA is utilized in the self-training process to provide proper hard examples for promoting the training process.

\subsection{Main results and Comparison with SOTA}
\vspace{-0.1cm}
\noindent
\textbf{Main results of our ST3D.}~
As shown in Table~\ref{tab:SOTAcomparison}, we compare the performance of our ST3D with Source Only, SN~\cite{wang2020train} and Oracle. 
Since SN employs extra statistical supervision on  the target domain, we compare our method with other approaches in terms of two settings, the Unsupervised DA (UDA) and Weakly-supervised DA setting (with target domain size statistics).
 
For the UDA setting, our method outperforms the Source Only baseline on all evaluated UDA settings. Specifically, without leveraging the target domain size statistics, we improve the performance on Waymo $\rightarrow$ KITTI and nuScenes $\rightarrow$ KITTI tasks by a large margin of around 34\% $\sim$ 43\% in $\text{AP}_{\text{3D}}$, which largely closes the performance gap between Source Only and Oracle. 
Furthermore, when transferring Waymo models to other domains that have full ring view annotations for evaluation (\ie, Waymo $\rightarrow$ nuSenses and Waymo $\rightarrow$ Lyft \footnote{Lyft dataset is constructed with different label rules from the other 3 datasets which enlarges the domain gaps and we will detail this in the supplementary materials}), 
our ST3D also attains a considerable performance gain which closes the Oracle and Source Only performance gap by up to $33.93\%$ on SECOND-IoU and $15.20\%$ on PV-RCNN. 
These encouraging results validate that our method can effectively adapt 3D object detectors trained on the source domain to the target domain and perform generally well on different detection architectures.


For the weakly-supervised DA setting, we equip our ST3D with the SN~\cite{wang2020train} (denoted as ST3D~(w/SN)) to obtain the pre-trained detector.
We observe that our ST3D approach and SN can work collaboratively to further boost the performance on Waymo $\rightarrow$ KITTI where ST3D improves SN by 14\% (SECOND-IoU) and 13\% (PV-RCNN) in $\text{AP}_\text{3D}$.
Notably, our ST3D (w/ SN) performs on par with the fully supervised 3D detector on this setting as shown in Table~\ref{tab:SOTAcomparison}.
Moreover, our approach with SECOND-IoU obtains over 40\% $\text{AP}_\text{3D}$ improvement on the nuScenes $\rightarrow$ KITTI setting compared with SN. 
For Waymo $\rightarrow$ nuScenes and Waymo $\rightarrow$ Lyft tasks, despite performance gains are still obtained compared to SN, only minor performance gains or even performance degradation are observed compared to our UDA setting ST3D due to the minor domain shifts in object size. In contrast, our ST3D still demonstrates consistent improvements on these settings.

We also observe that it is hard to adapt detectors from the point clouds with more LiDAR beams (\eg Waymo) to the point clouds with fewer LiDAR beams (\eg NuScenes), while the opposite adaptation is relatively easy as shown in Table~\ref{tab:SOTAcomparison} nuScenes $\rightarrow$ KITTI. It demonstrates that the point density of target domain is more important than the point density of source domain, and our ST3D could effectively improve the performance on target domain even with a relatively worse pre-trained detector on source domain.

\subsection{Ablation Studies}
In this section, we conduct extensive ablation experiments to investigate the individual components of our ST3D. All experiments are conducted with the 3D detector SECOND-IoU on the task of Waymo $\rightarrow$ KITTI.

\begin{table}
    \centering
    \begin{small}
        \begin{tabular}{l|c}
            \bottomrule[1pt]
            Method  & AP$_{\text{BEV}}$ / AP$_{\text{3D}}$ \\
            \hline
            (a) Source Only & 67.64 / 27.48 \\
            (b) Random Object Scaling (ROS) &  78.07 / 54.67  \\
            (c) SN  & 78.96 / 59.20 \\
            \hline
            (d) ST3D (w/o ROS) & 75.54 / 34.76 \\
            (e) ST3D (w/ ROS)  &  82.19 / 61.83 \\
            (f) ST3D (w/ SN) & \textbf{85.83} / \textbf{73.37} \\
            \toprule[0.8pt]
        \end{tabular}
    \end{small}
    \caption{Effectiveness analysis of Random Object Scaling.}
    \vspace{-0.5cm}
    \label{tab:random_object_size}
\end{table}

\noindent
\textbf{Random Object Scaling.}~
As mentioned in Sec.~\ref{sec:pretrain}, 
by employing our random object scaling for pre-training, the detectors could be more robust to the variations of object size in different domains.
Table~\ref{tab:random_object_size} (a), (b), (c) show that our unsupervised ROS improves the performance by around 27.2\% in $\text{AP}_\text{3D}$ and is only 4.5\% lower than the weakly-supervised SN method.
Furthermore, as shown in Table~\ref{tab:random_object_size} (d), (e), 
the ROS pre-trained model also greatly benefits the subsequent self-training process.
We also observe that there still exists a gap between the performance of ST3D (w/ ROS) and ST3D (w/ SN)  in $\text{AP}_\text{3D}$, potentially due to that the KITTI dataset has a larger domain gap over object size compared with other datasets, and in this situation, the weakly supervised SN could provide more accurate object size information than our fully unsupervised ROS.

\begin{table}
    \centering
    \begin{small}
    	\scalebox{0.92}{
        \begin{tabular}{l|c}
            \bottomrule[1pt]
            Method  & AP$_{\text{BEV}}$ / AP$_{\text{3D}}$ \\
            \hline
            SN  (baseline)  & 78.96 / 59.20 \\
            \hline
            ST (w/ SN) &  79.74 / 65.88 \\
            ST (w/ SN) + Triplet       & 79.81 / 67.39  \\
            ST (w/ SN) + Triplet + QAC  & 83.76 / 70.64   \\
            ST (w/ SN) + Triplet + QAC + MEV-C & 85.35 / 72.52 \\
            ST (w/ SN) + Triplet + QAC + MEV-C + CDA  & \textbf{85.83} / \textbf{73.37} \\
            \toprule[0.8pt]
        \end{tabular}
    }
    \end{small}
    \caption{Component ablation studies. \textbf{ST} represents naive self-training. \textbf{Triplet} means the triplet box partition. \textbf{QAC} indicates the quality-aware criterion. \textbf{MEV-C} is consistency memory ensemble-and-voting. \textbf{CDA} means curriculum data augmentation.}
    \vspace{-0.1cm}
    \label{tab:component_analysis}
\end{table}

\noindent
\textbf{Component Analysis in Self-training.}~
As demonstrated in Table~\ref{tab:component_analysis}, we investigate the effectiveness of our individual components. Our ST3D (last line) outperforms the SN baseline and naive ST (w/ SN) by around 14.2\% and 7.5\% in $\text{AP}_{\text{3D}}$. Specifically, on the pseudo label generation stage, Triplet box partition and quality-aware IoU criterion provide around 1.5\% and 3.3\% performance gains on $\text{AP}_{\text{3D}}$, respectively. MEV-C and CDA separately further yield around 1.9\% and 0.9\% improvements, respectively.

\begin{table}
    \definecolor{Gray}{gray}{0.9}
    \newcolumntype{a}{>{\columncolor{Gray}}c}
    \centering
    \vspace{-0.1cm}
    \begin{small}
        \begin{tabular}{lac|acc}
            \bottomrule[1pt]
            $T_{\text{neg}}$ & $T_{\text{pos}}$ & AP$_{\text{BEV}}$ / AP$_{\text{3D}}$ & $T_{\text{neg}}$ & $T_{\text{pos}}$ & AP$_{\text{BEV}}$ / AP$_{\text{3D}}$ \\
            \hline
            0.20 & 0.60  & \textbf{86.44} / 72.23 & 0.25 & 0.25 & 83.06 / 67.97 \\
            0.25 & 0.60 & 85.83 / \textbf{73.37} & 0.25 & 0.30 & 83.21 / 69.51 \\
            0.30 & 0.60 & 85.30 / 72.73 & 0.25 & 0.40 & 83.69 / 69.98 \\
            0.40 & 0.60 & 84.59 / 72.25 & 0.25 & 0.50 & 84.30 / 70.17 \\
            0.50 & 0.60 & 84.96 / 72.11 & 0.25 & 0.60 & \textbf{85.83} / \textbf{73.37} \\
            0.60 & 0.60 & 83.66 / 70.10 & 0.25 & 0.70 & 76.81 / 66.23 \\
            \toprule[0.8pt]
        \end{tabular}
    \end{small}
    \caption{Sensitivity analysis for [$T_{\text{neg}}$, $T_{\text{pos}}$] of triplet box partition.}
    \vspace{-0.4cm}
    \label{tab:triplet}
\end{table}

\noindent
\textbf{Sensitivity Analysis of Triplet Box Partition.}~ 
In this part, we investigate the importance of the ignore margin [$T_{\text{pos}}$, $T_{\text{neg}}$] for our triplet box partition. As shown in Table~\ref{tab:triplet}, without triplet box partition (i.e., $T_{\text{pos}}$ = $T_{\text{neg}}$), our ST3D drops by 3.3\% and 5.4\% for $T_{\text{pos}}$ = $T_{\text{neg}}$ = 0.6 and 0.25 respectively. Furthermore, our method is more sensitive to $T_{\text{pos}}$ than $T_{\text{neg}}$. Lower $T_{\text{pos}}$ could introduce excessive noisy labels while higher $T_{\text{pos}}$ gives rise to a small number of positive examples that harm the self-training process.

\noindent
\textbf{Analysis of Memory Ensemble and Voting.}
\label{abl:mev}
As shown in Table~\ref{tab:abl_memory_ensemble}, we  further investigate the memory ensemble and memory voting  schemes for updating memory bank and generating pseudo labels. On the one hand, we propose the other two memory ensemble strategies including NMS ensemble and bipartite ensemble, which use NMS and bipartite matching separately.
For the comparison of different memory ensemble variants, ME-N and ME-C achieve similar performance and outperform 0.8\% $\sim$ 1\% than ME-B in terms of 3D AP.
For the paired box merging strategy in the memory ensemble stage, we compare two merging approaches max score and weighted average, where max score obtains a 1.3\% performance gain than weighted average. This validates our analysis in Sec.~\ref{sec:memoryensemble} that the weighted average strategy may generate inappropriate pseudo labels when matched boxes have very different heading angles.

On the other hand, without memory voting, the performance drops over 2.4\% since the unmatched boxes along different memories could not be well handled. Our memory voting strategy could robustly mine high-quality boxes and discard low-quality boxes.

\begin{table}
	\centering
	\scalebox{0.95}{
		\setlength\tabcolsep{1pt}
		\begin{small}
			\begin{tabular}{c|c|c|c}
				\bottomrule[1pt]
				Method  & Memory Voting & Merge & AP$_{\text{BEV}}$ / AP$_{\text{3D}}$ \\
				\hline
				ST3D (w/ ME-N) & $\surd$ & Max & \textbf{85.93} / 73.17 \\
				ST3D (w/ ME-B) & $\surd$ & Max & 85.65 / 72.37 \\
				\hline
				\multirow{4}{*}{ST3D (w/ ME-C)} & $\surd$ & Max & 85.83 / \textbf{73.37} \\
				& $\surd$ & Avg & 84.08 / 72.07 \\
				\cline{2-4}
				& $\times$  & Max & 84.23 / 70.86 \\
				& $\times$  & Avg & 83.92 / 70.96 \\ 
				\toprule[0.8pt]
			\end{tabular}
		\end{small}
	}
	\caption{Ablation study of memory ensemble (different variants and merge strategies for matched boxes) and memory voting. We denote three memory ensemble variants: consistency, NMS and bipartite as ME-C, ME-N, ME-B separately.}
	\vspace{-0.1cm}
	\label{tab:abl_memory_ensemble}
\end{table}

\begin{table}
    \centering
    \scalebox{0.92}{
        \begin{tabular}{c|cc|c|c}
            \bottomrule[1pt]
            Method  & World & Object & Intensity & AP$_{\text{BEV}}$ / AP$_{\text{3D}}$ \\
            \hline
            \multirow{6}{*}{ST3D} &$\times$  &$\times$  & - & 83.31 / 66.73 \\
             & $\surd$ & $\times$ & Normal & 84.47 / 70.60 \\
             &$\times$  & $\surd$ & Normal & 81.81 / 67.91 \\
             & $\surd$ & $\surd$ & Normal & 85.35 / 72.52 \\
            \cline{2-5} 
            & $\surd$ & $\surd$ & Strong & 84.84 / 72.23 \\
            & $\surd$ & $\surd$ & Curriculum & \textbf{85.83} / \textbf{73.37} \\
            \toprule[0.8pt]
        \end{tabular}
    }
    \caption{Analysis of data augmentation type and intensity.}
    \vspace{-0.2cm}
    \label{tab:abl_aug}
\end{table}


\noindent
\textbf{Data Augmentation Analysis.}
As shown in Table~\ref{tab:abl_aug}, we also investigate the effects of data augmentation in the self-training pipeline, where both the type (world-level and object-level) and the intensity of augmentation are explored. 
We observe that without any data augmentation, ST3D suffers from over 6.6\% performance degradation. Both world-level and object-level augmentation provide improvements and their combination can further boost the performance. 
When it comes to the intensity of data augmentation, compared to the normal intensity, stronger data augmentation magnitude confuses the deep learner and slightly drops performance while our CDA can bring around 0.9\% gains.

\begin{figure}[h]
    \centering
    \includegraphics[width=1\linewidth]{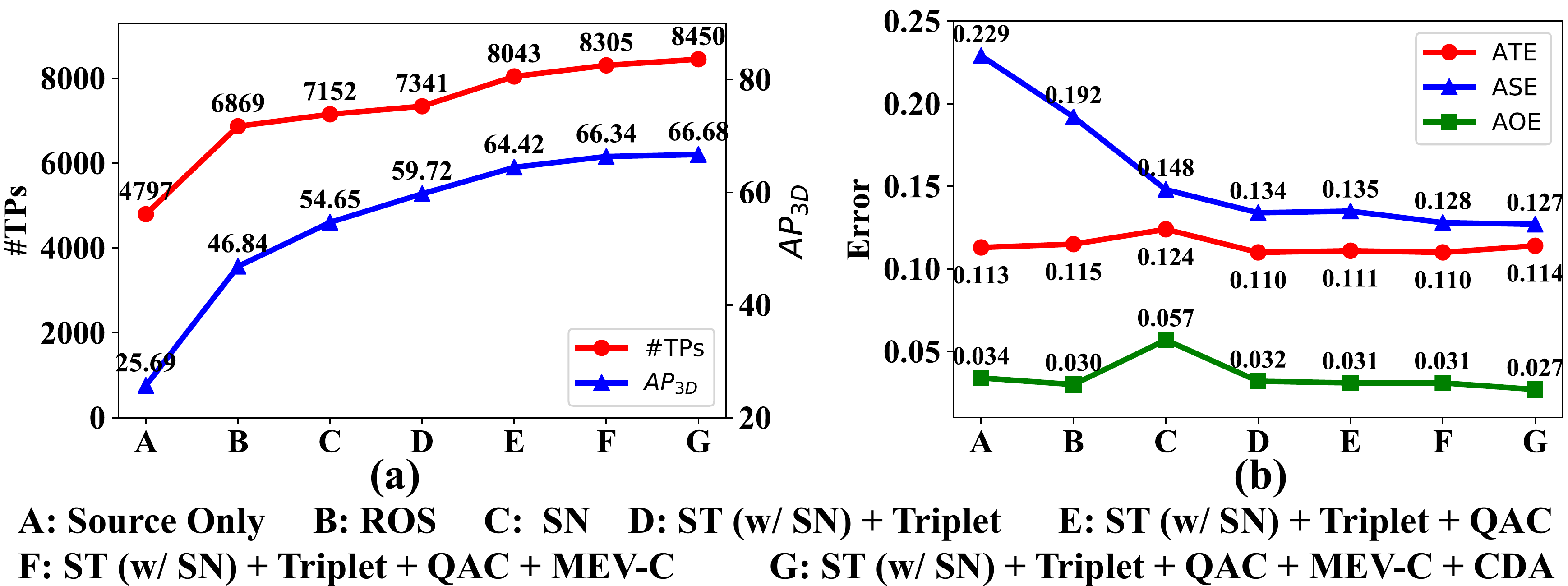}
    \vspace{-0.4cm}
    \caption{Quality of pseudo labels on KITTI training set.}
    \label{fig:quality}
    \vspace{-0.3cm}
\end{figure}

\noindent
\textbf{Quality of Pseudo Labels.}
To directly investigate how each component contribute to the quality of pseudo labels, we utilize $AP_{\text{3D}}$ and \#TPs to assess the correctness of pseudo labels. Besides,  \textbf{ATE}, \textbf{ASE} and \textbf{AOE} are to measure the translation, scale and orientation errors (refer to nuScenes toolkit~\cite{caesar2020nuscenes}). As shown in  Figure~\ref{fig:quality}, 
ROS mitigates domain differences in object size distributions and hence largely reduces ASE; with Triplet, QAC and MEV, our method generates accurate and stable pseudo labels, localizing more \#TPs with fewer errors; and
CDA overcomes overfitting and reduces both ASE and AOE.

\section{Conclusion}
We have presented ST3D -- a redesigned self-training pipeline -- for unsupervised domain adaptive 3D object detection from point clouds. ST3D involves random object scaling, a quality-aware triplet memory bank, and curriculum data augmentation to address fundamental challenges stemming from the self-training on 3D object detection. Experiments demonstrate that ST3D substantially advance the state of the art. 
Our future work will be to extend our method to other UDA tasks on image and video data.



{\small
\bibliographystyle{ieee_fullname}
\bibliography{egbib}
}

\clearpage
\appendix

\renewcommand{\thesection}{S\arabic{section}}
\renewcommand{\thetable}{S\arabic{table}}
\renewcommand{\thefigure}{S\arabic{figure}}

\centerline{\large{\textbf{Outline}}}
In this supplementary file, we provide more details and visualizations omitted in our main paper due to 8-pages limits on paper length:
\vspace{-0.15cm}
\begin{itemize}
    \setlength{\itemsep}{1.0pt}
    \setlength{\parsep}{1.0pt}
    \setlength{\parskip}{1.0pt}
    \item Sec.~\ref{sec:Datasets}: Dataset details for our domain adaptation tasks.
    \item Sec.~\ref{sec:lyft_annotation_problems}: Analysis of domain difference and systematic bias on pseudo labels .
    \item Sec.~\ref{sec:implementation_details}: Implementation details for SECOND-IoU and other memory ensemble variants.
    \item Sec.~\ref{sec:iou0.5_reuslts}: More experimental results with IoU threshold at 0.5.
    \item Sec.~\ref{sec:extra_ablation_studies}: Additional ablation studies.
    \item Sec.~\ref{sec:qualitative_results}: Qualitative results.
    \item Sec.~\ref{sec:extra_tasks}: Experiments on other adaptation tasks.
\end{itemize}
\section{Dataset Overview} \label{sec:Datasets}
\begin{figure}[htbp]
	\centering
	\includegraphics[width=\linewidth]{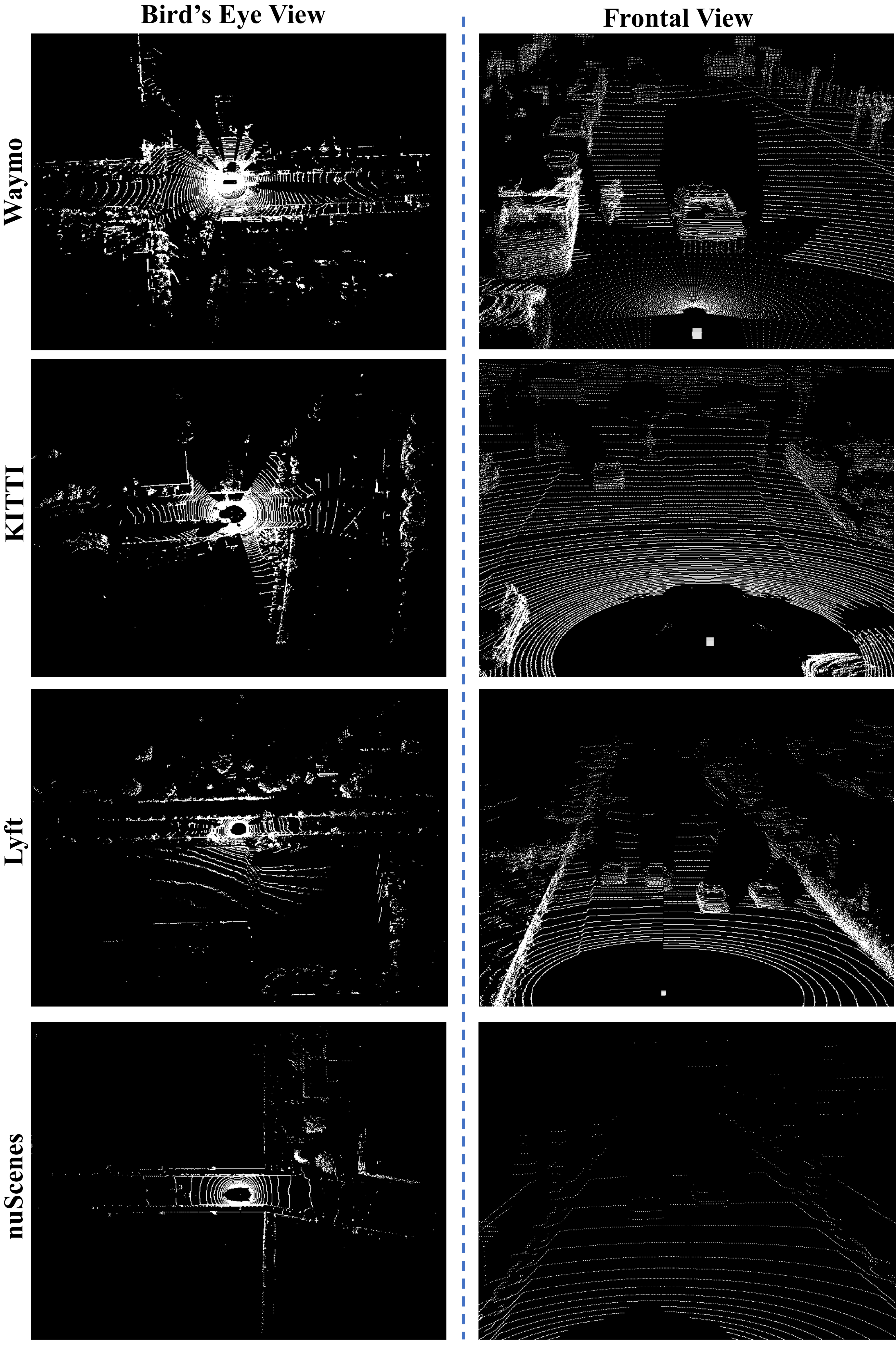}
	\caption{Visualization of bird's eye views (left) and frontal views (right) for different datasets: Waymo~\cite{sun2020scalability}, KITTI~\cite{Geiger2012KITTI}, Lyft~\cite{lyft2019} and nuScenes~\cite{caesar2020nuscenes}. nuScenes has obviously sparse point clouds than other three datasets since it is only collected by 32-beam LiDAR. Even Waymo, KITTI and Lyft all utilize 64-beam LiDARs, Waymo is denser than KITTI and Lyft and its beams are clustered in the medium of vertical angles.}
    \label{fig:sup_data_comp}
    \vspace{-0.3cm}
\end{figure}

We compare four LiDAR 3D object detection datasets as shown in Table~\ref{tab:sup_dataset_infos}. They are different in LiDAR type, beam angles, point cloud density, size, and locations for data collection. Visual illustrations in Figure~\ref{fig:sup_data_comp} obviously show the different patterns of LiDAR point clouds in terms of distribution and density. Even for data from LiDARs with same beams (Waymo, KITTI, and Lyft in Figure~\ref{fig:sup_data_comp}), point clouds are also different in the range, vertical, and horizontal distributions. For instance, Waymo not only utilizes a small horizontal azimuth of LiDAR, but also clusters LIDAR beams in the medium of vertical angles (see Figure~\ref{fig:sup_data_comp}). Both these LiDAR setups lead to denser point clouds in the collected data (see \textit{\# points per scene} in Table~\ref{tab:sup_dataset_infos}).

We conduct experiments on domain adaptations from the label-rich domain to label-insufficient domains ({\ie} Waymo $\rightarrow$ KITTI, Waymo $\rightarrow$ Lyft, Waymo $\rightarrow$ nuScenes) and the more challenging domain adaptations across domains with the different number of LiDAR beams ({\ie} Waymo $\rightarrow$ nuScenes and nuScenes $\rightarrow$ KITTI).
On all evaluated settings, our approach improves the baseline method and outperforms the existing approach by a significant margin, demonstrating the efficacy of the proposed approach.

\begin{table*}
    \centering
    \scalebox{0.88}{
    \begin{tabular}{l|c|c|c|c|c|c}
        \bottomrule[1pt]
        Dataset  & LiDAR Type & Beam Angles & \# Points Per Scene$^{\dagger}$ &  \# Training Frames & \# Validation Frames & Location \\
        \hline
        Waymo \cite{sun2020scalability} & 64-beam & [-18.0$^{\circ}$, 2.0$^{\circ}$]$^{*}$ & 160,139 & 158,081 & 39,987 & USA \\
        \hline
        KITTI \cite{Geiger2012KITTI} & 64-beam & [-23.6$^{\circ}$, 3.2$^{\circ}$] & 118,624 & 3,712 & 3,769 & Germany \\
        \hline
        Lyft \cite{lyft2019} & 64-beam & [-29.0$^{\circ}$, 5.0$^{\circ}$]$^{*}$ & 69,175 & 18,900 & 3,780 & USA \\
        \hline
        nuScenes \cite{caesar2020nuscenes} & 32-beam & [-30.0$^{\circ}$, 10.0$^{\circ}$] & 24,966 & 28,130 & 6,019 & USA and Singapore \\
        \toprule[0.8pt]
    \end{tabular}
    }
    \caption{Dataset overview. Notice that we use \textbf{version 1.0} of Waymo Open Dataset. * indicates we obtain the information from \cite{wang2020train}. ${\dagger}$ means that we count this statistical information only on the validation set.}
    \label{tab:sup_dataset_infos}
\end{table*}

\begin{figure*}[htbp]
	\centering
	\includegraphics[width=\linewidth]{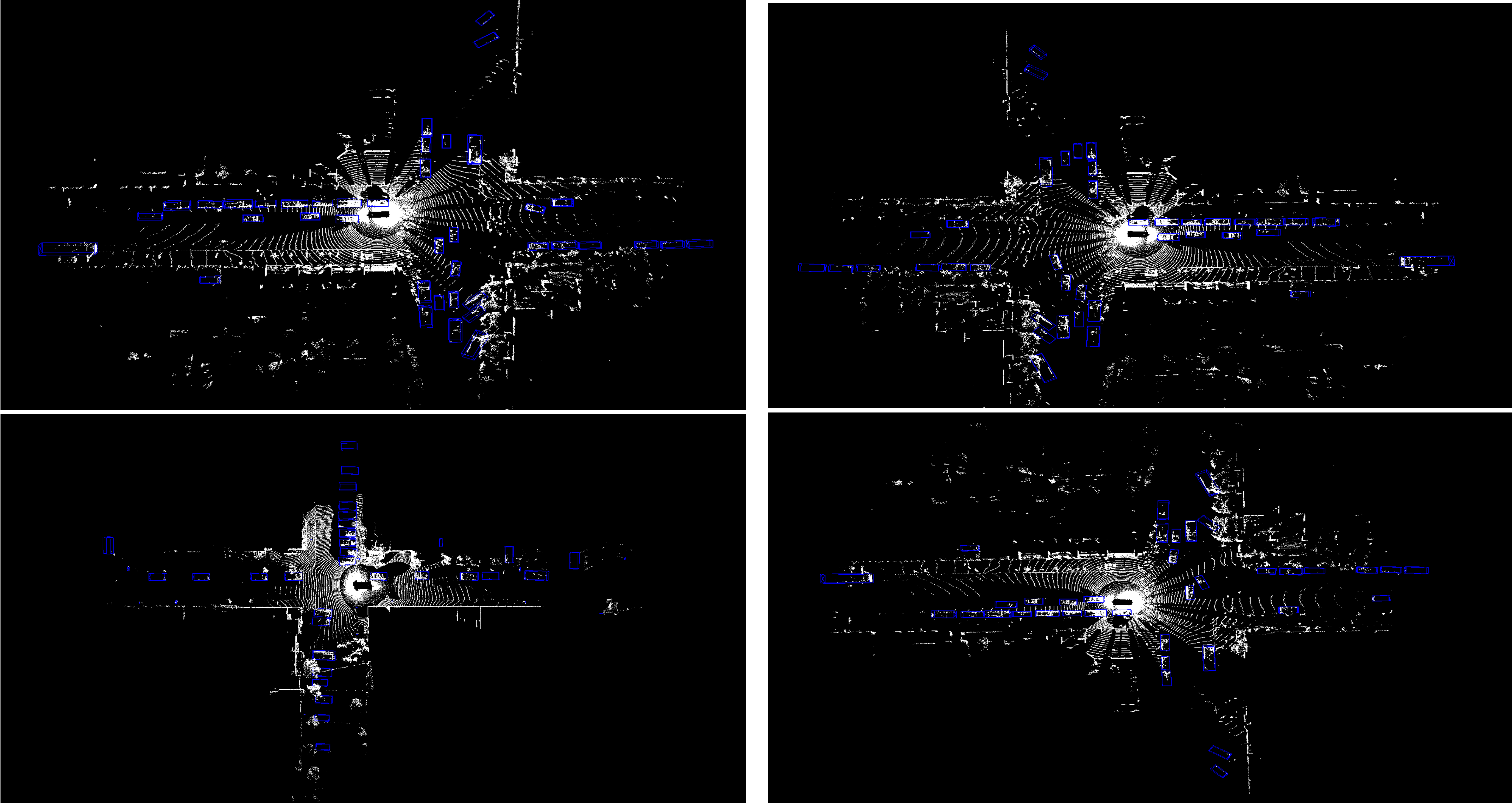}
	\caption{Examples of Waymo scenes. The blue boxes are ground-truth bounding boxes.}
    \label{fig:sup_waymo_result}
    \vspace{-0.2cm}
\end{figure*}

\begin{figure*}[htbp]
	\centering
	\includegraphics[width=\linewidth]{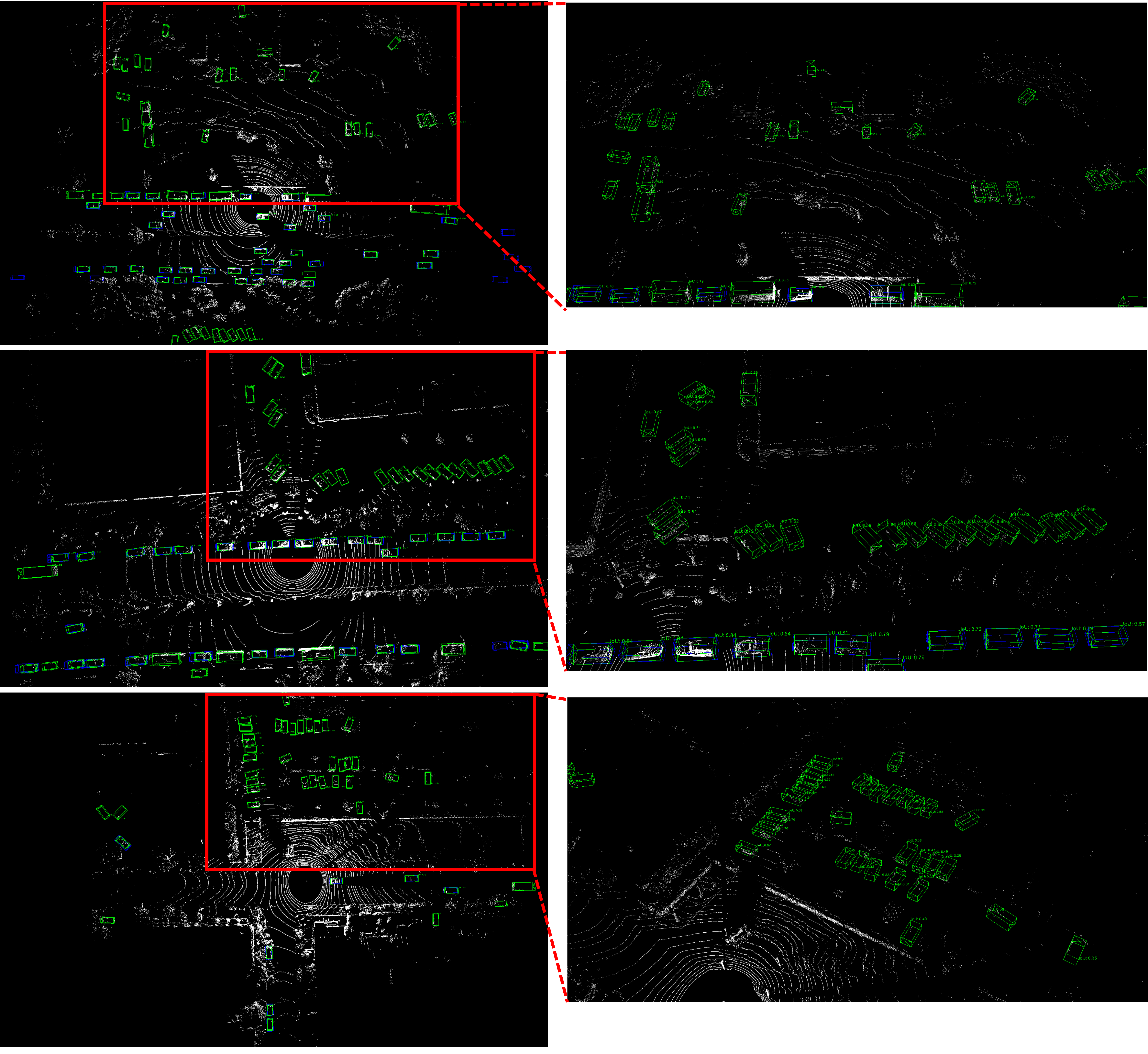}
	\caption{Examples to show the annotation gap between Lyft and Waymo. The green boxes are prediction results from the Waymo pre-trained model while the blue boxes are Lyft annotated boxes.}
    \label{fig:sup_lyft_annos}
\end{figure*}

\section{Domain Difference and Systematic bias}
\label{sec:lyft_annotation_problems}
\subsection{Lyft Annotation Discrepancies} 
The Lyft~\cite{lyft2019} dataset is constructed by a labeling protocol different from the other three datasets, {\ie} the Lyft dataset does not annotate objects on both sides of the road. For instance, we observe that the objects on the main branch of the road (\textit{w.r.t} the ego car) are most likely annotated, while many objects on both sides might not be annotated. 
Visual illustrations of the annotated bounding boxes are shown in Fig.~\ref{fig:sup_waymo_result} for the Waymo dataset (\textcolor{blue}{blue} boxes) and Fig.~\ref{fig:sup_lyft_annos} (\textcolor{blue}{blue} boxes) for the Lyft dataset. 

The differences in annotation protocols will have a negative influence on the evaluation of domain adaptation results. 
When we use the pre-trained model on the Waymo dataset to evaluate data from the Lyft scenes, our model correctly predicts the cars on two sides of the road (see \textcolor{green}{green} boxes in Fig~\ref{fig:sup_lyft_annos}), which, however, are not annotated by the Lyft dataset (see \textcolor{blue}{blue} boxes in Fig.~\ref{fig:sup_lyft_annos}). This makes it hard to evaluate the actual performance boost with the proposed domain adaptation method. We believe that our method can obtain a further performance boost if the results are properly evaluated.


\subsection{Analysis of Domain Discrepancy}
We conclude that the domain gap mainly lies in two folds: ($i$) content gap (\eg object size) caused by different data-capture locations; ($ii$) point distribution gap caused by different LiDAR beams. Self-training explicitly closes the domain gap by reformulating the UDA problem as a target domain supervised problem with pseudo labels, where better pseudo labels provide better performance. 

\subsection{Systematic Bias on Pseudo Labels}
An important systematic bias on pseudo labels is \textit{Annotation style bias} due to different annotation rules such as how to annotate (tightness of bounding boxes) and which to annotate (See Sec.~\textcolor{red}{S2.1} in Suppl.). This will make pseudo labels biased toward the source domain labeling rules, different from target domain GT.

\section{Implementation details} \label{sec:implementation_details}
In this section, we give more implementation details in constructing our adaptation tasks.
Further, we illustrate the component selection of the oracle model, the IoU head of SECOND~\cite{yan2018second} as well as the other two memory ensemble variants: NMS ensemble and bipartite ensemble.

\subsection{Parameter setups}
We typically pre-train the detector for 30 epochs on Waymo and then train 30 epochs for self-training to converge on Waymo $\rightarrow$ KITTI setting. Besides, we update pseudo labels every two epoch. The scaling range of ROS is [0.75, 1.1], ensuring a reasonable scaled car size. For the QTMB, the two thresholds $T_{\text{neg}}$ and $T_{\text{pos}}$ of triplet box partition are 0.25 and 0.6, respectively. As for CDA, we split the total self-training epochs into six stages (i.e., epochs [0, 5), [5, 10), [10, 15), [15, 20), [20, 25), [25, 30)). More detailed parameter setups could be found in our released code.

\subsection{Details of Voxel Size and GT Sampling for Oracle Model.}
Here, we provide more details on the voxel size for SECOND-IOU and the GT sampling strategy for training.
\paragraph{Voxel Size.} We derive our Oracle model with voxel size [0.10m, 0.10m, 0.15m] rather than [0.05m, 0.05m, 0.15m]
To be noted, we adopt this setting in \textbf{all} experiments including our pre-trained model and self-training pipeline for a fair evaluation. The reason why we adopt this setting is that all our models are trained with the ring view (about $150m \times 150m$) which will take too much GPU memory if the voxel size is set to [0.05m, 0.05m, 0.15m] (we can only set batch size as 1 for SECOND-IoU and totally fail to run PV-RCNN with such voxel size). We use NVIDIA GTX 1080Ti with 11G GPU memory for all experiments and adopt voxel size [0.10m, 0.10m, 0.15m] to 
achieve the best trade-off between memory and realization in various settings as well as frameworks.

\paragraph{GT Sampling.} We do not adopt the GT sampling data augmentation for all settings for fair comparisons. The reason is that it is unaffordable for the iterative self-training pipeline to use GT sampling data augmentation since it requires frequently generating a new GT database with updated pseudo labels, which produces a large computation cost (leveraging GT sampling for self-training takes more than 3$\times$ training time).

\begin{table}
	\centering
	\scalebox{0.83}{
		\begin{small}
			\begin{tabular}{l|c|c|c}
				\bottomrule[1pt]
				Method  & Voxel Size & GT Sampling  & AP$_{\text{BEV}}$ / AP$_{\text{3D}}$ \\
				\hline
                \multirow{4}{*}{Oracle (Ours)}  & [0.10m, 0.10m, 0.15m] &  &  83.29 / 73.45 \\
                \cline{2-4}
                & [0.05m, 0.05m, 0.15m] &  & 85.99 / 76.53 \\
				 & [0.10m, 0.10m, 0.15m] & $\surd$ & 88.08 / 81.52 \\
				 & [0.05m, 0.05m, 0.15m] &  $\surd$ & \textbf{88.56} / \textbf{81.87} \\
                \hline
                Oracle (SN~\cite{wang2020train}) & - & unknown & 80.60 / 68.90 \\
				\toprule[0.8pt]
			\end{tabular}
		\end{small}
	}
    \caption{Comparison of different setting (voxel size and GT sampling) for our Oracle model based on SECOND-IoU. We also compare them with the Oracle performance release in the SN~\cite{wang2020train} based on PointRCNN. The reported AP results are evaluated on the moderate difficulty of the car category of the KITTI validation set at IoU threshold 0.7.}
	\label{tab:sup_abl_fully}
\end{table}

\paragraph{More Analysis.} 
Here, we show the oracle results trained with voxel size [0.05m, 0.05m, 0.15m] and GT sampling data augmentations. The results are listed in Table~\ref{tab:sup_abl_fully}. 
Though our model performance presented in Table 1 in our paper is obtained using a sub-optimal setup for memory and computational efficiency, our adaptation results are still competitive in comparison with results in Table~\ref{tab:sup_abl_fully}. 
Furthermore, employing PointRCNN as the framework, Oracle results in SN~\cite{wang2020train} even has 4.55\% performance gap to our sub-optimal Oracle model.
It is noteworthy that, the development of the ST3D model is orthogonal with the above modifications, and ST3D could also benefit from these training modifications and further boost the performance.

We would like to highlight that our focus in this paper is to demonstrate the effectiveness of ST3D without adopting various training tricks in 3D object detection. And we believe the presented comparisons in the main paper are fair and could assess the actual progress made by our ST3D pipeline.

\subsection{SECOND-IoU}
Given the object proposals from the RPN head in the original SECOND network, we extract the proposal features from 2D BEV features using the rotated RoI-align operation~\cite{he2017mask}. Then, taking the extracted features as inputs, we adopt two fully connected layers with ReLU nonlinearity~\cite{agarap2018deep} and batch normalization~\cite{ioffe2015batch} to regress the IoU between RoIs and their corresponding ground-truths (or pseudo boxes) with sigmoid nonlinearity. During training, we do not back-propagate the gradient from our IoU head $\mathcal{L}_{\text{iou}}$ to our backbone network. We observe the attached IoU branch could also boost the performance of the baseline SECOND model, namely SECOND-IoU, if the IoU prediction score is used for NMS.


\subsection{Other Memory Ensemble Variants}
\paragraph{NMS ensemble} is an intuitive solution to match and merge boxes based on the IoU between two boxes. It directly removes matched boxes with lower confidence scores.
Specifically, we concatenate historical pseudo labels and current proxy-pseudo labels to $[\tilde{M}^{t}_i]_{k} = \{[M^{t}_i]_{k-1}, [\hat{L}^t_i]_k\}$ as well as their corresponding confidence scores to $\tilde{u}^k_i = \{u^{k-n}_i, u^k_i \}$ for each target sample $P^t_i$. Then, we obtain the final pseudo boxes $[M^{t}_i]_{k}$ and corresponding confidence score $u^k_i$ by applying NMS with a IoU threshold $0.1$ as
\begin{equation}
    [M^{t}_i]_{k}, u^k_i = \text{NMS}([\tilde{M}^{t}_i]_{k},\  \tilde{u}^k_i).
\end{equation}

\paragraph{Bipartite ensemble} employs optimal bipartite matching to pair historical pseudo labels $[M_i^{t}]_{k-1}$ and current proxy-pseudo labels $[\hat{L}^t_i]_k$ and then follow consistency ensemble to process matched pairs. Concretely, we assume that there are $n_m$ and $n_l$ boxes for $[M_i^{t}]_{k-1}$ and $[\hat{L}^t_i]_k$ separately. Then, we search a permutation of $n_m$ elements $\sigma \in \mathfrak{S}_{n_m}$ with the lowest cost as
\begin{equation}
    \hat{\sigma}=\underset{\sigma \in \mathfrak{S}_{n_m}}{\arg \min } \sum_{j}^{n_m} \mathcal{L}_{\operatorname{match}}\left(b_{j}, b_{\sigma(j)}\right),
\end{equation}
where the matching cost $\mathcal{L}_{\operatorname{match}}$ is the $-\text{IoU}$ between the matched boxes. Notice that the matched box pairs with IoU lower than 0.1 would still be regarded as unmatched.

\section{Experimental Results with IoU = 0.5} \label{sec:iou0.5_reuslts}
In this section, we report the AP$_\text{BEV}$ and AP$_\text{3D}$ with the IoU threshold \textbf{0.5} as a supplement to the experimental results in our main submission. The results are shown in Table~\ref{tab:sup_random_object_size}, \ref{tab:sup_component_analysis}, \ref{tab:sup_triplet}, \ref{tab:sup_abl_memory_ensemble} and \ref{tab:sup_abl_aug}, \ref{tab:sup_SOTAcomparison}. To be noted, IoU threshold \textbf{0.7} is a more strict criterion and widely adopted to assess 3D object detection models for the ``car'' category~\cite{shi2019points,yan2018second,shi2020pv,wang2020train}.

\section{Extra Ablation Studies} \label{sec:extra_ablation_studies}
In this section, we present more ablation experiments and analysis. 
All experiments are conducted with the 3D detector SECOND-IoU on the adaptation setting of Waymo $\rightarrow$ KITTI. Our reported AP results are evaluated on the moderate difficulty of the car category of the KITTI dataset.

\begin{table}[htpb]
	\centering
	\scalebox{0.87}{
		\begin{small}
			\begin{tabular}{l|c|c|c|c}
				\bottomrule[1pt]
				Method  & Framework & Sequence & $\text{AP}_{\text{3D}}$ & Closed Gap \\
				\hline
				Source Only & PointRCNN & unknown & 21.9  & - \\
				SF-UDA$^{3D}$~\cite{saltori2020sf} & PointRCNN & $\surd$ & \textbf{54.5}  & 56.0\% \\
				Oracle & PointRCNN &  & 80.1 &  - \\
				\hline
				Source Only & SECOND-IoU &  & 17.9  & - \\
				ST3D & SECOND-IoU &  & 54.1 & \textbf{65.1\%} \\
				Oracle & SECOND-IoU &  & 73.5  & - \\
				\toprule[0.8pt]
			\end{tabular}
		\end{small}
	}
    \caption{Comparison with SF-UDA$^{3D}$ on nuScenes $\rightarrow$ KITTI.}
	\label{tab:sup_SOTA_con}
\end{table}

\paragraph{Compared with the Contemporary SOTA.}~
As shown in Table~\ref{tab:sup_SOTA_con}, SF-UDA$^{3D}$ is a contemporary work that leverages the consistency of \textbf{temporal information} along with the point cloud sequences to address the domain shift on 3D object detection. 
By using \textbf{only the single-frame} point cloud as input, our ST3D achieves similar performance while being much closer to the fully-supervised oracle results.

\begin{table}
	\centering
	\scalebox{0.95}{
		\begin{small}
			\begin{tabular}{c|c|c|c}
				\bottomrule[1pt]
				Method  & Confidence  & AP$_{\text{BEV}}$ / AP$_{\text{3D}}$ & Gain \\
				\hline
                \multirow{2}{*}{SN} & Classification & 77.68 / 57.08 & - \\
                 & IoU & 78.96 / 59.20 &  1.28 / 2.12 \\
                \hline
                \multirow{2}{*}{ST3D (w/ SN)} & Classification & 82.21 / 69.58 & - \\
                 & IoU & \textbf{85.83} / \textbf{73.37} & \textbf{3.62} / \textbf{3.79} \\
                \hline
                \multirow{2}{*}{Oracle} & Classification & 84.48 / 73.01 & \\
                 & IoU & 83.29 / 73.45 & -0.99 / 0.44 \\
				\toprule[0.8pt]
			\end{tabular}
		\end{small}
	}
    \caption{Comparison of different confidence criteria.}
    \vspace{-0.5cm}
	\label{tab:sup_cls_iou_compare}
\end{table}

\begin{figure*}
	\centering
	\includegraphics[width=\linewidth]{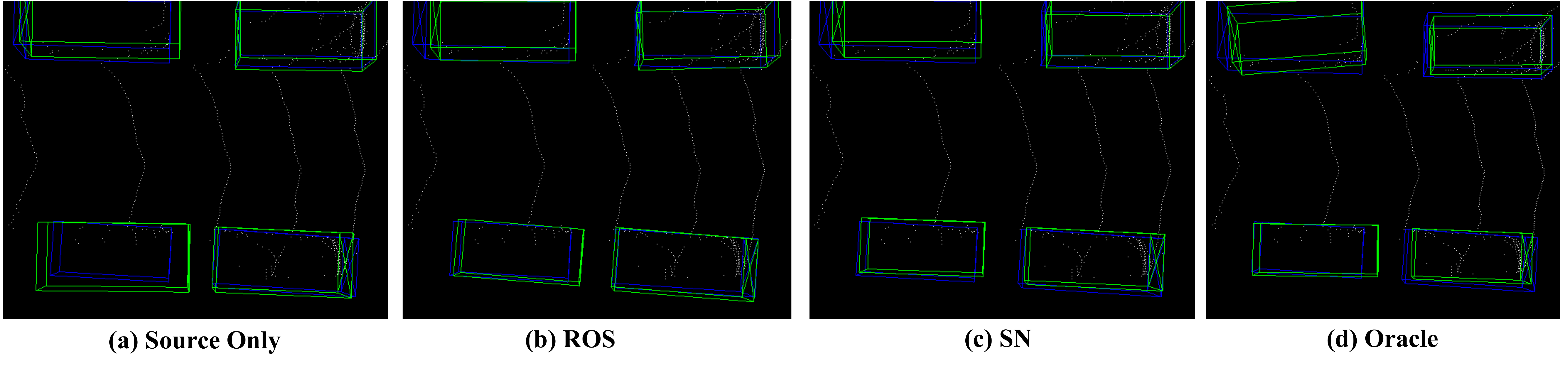}
	\caption{Comparison of ROS and SN to close object-size level domain gap on Waymo $\rightarrow$ KITTI. The green and blue bounding boxes are detector predictions and GTs, respectively. (a) Source Only: The detector is trained on Waymo without SN or ROS. (b) The detector is trained with ROS on Waymo. (c) The detector is trained with SN~\cite{wang2020train} on Waymo. (d) The detector is trained on KITTI.}
	\label{fig:sup_vis_ros}
\end{figure*}
\paragraph{Quality-aware Confidence Criterion.} Here, we investigate the influence of the IoU  confidence criterion on the pre-trained SN model, the self-training pipeline and the fully supervised oracle model, respectively. As illustrated in Table~\ref{tab:sup_cls_iou_compare}, the IoU score can bring performance improvements for all three settings in comparison with the classification score.
Specifically, the IoU confidence yields a 2.12\% gain for the SN model and a 0.44\% gain for the fully supervised oracle model in terms of AP$_{\text{3D}}$. More importantly,  our ST3D (w/ SN)  self-training pipeline could benefit more from the IoU criterion, obtaining as much as 3.79\% performance boost in items of AP$_{\text{3D}}$. 
This suggests that the IoU confidence criterion could facilitate the model to produce high-quality pseudo-labeled data, and ultimately lead to a much better 3D object detection model.


\section{Qualitative Results} \label{sec:qualitative_results}
\paragraph{Qualitative Results of Random Object Scaling.} 
We have compared the AP$_{\text{BEV}}$ and AP$_{\text{3D}}$ of our ROS with SN and Source Only model in the Table~2 of our main paper. Here, we provide qualitative results of the Source Only model, ROS, SN and Oracle for visual comparisons. As shown in Fig.~\ref{fig:sup_vis_ros}, the zoom-in regions in the left bottom box in each sub-figure shows that both SN and ROS can largely improve the localization accuracy of the pre-trained model while our ROS does not leverage extra statistical information on the target domain.


\begin{table}
    \centering
    \begin{small}
        \begin{tabular}{l|c}
            \bottomrule[1pt]
            Method  & AP$_{\text{BEV}}$ / AP$_{\text{3D}}$ \\
            \hline
            (a) Source Only & 91.52 / 89.94 \\
            (b) Random Object Scale (ROS) & 88.98 / 87.33  \\
            (c) SN  & 87.18 / 85.91 \\
            \hline
            (d) Ours (w/o ROS) & \textbf{93.68 / 92.50} \\
            (e) Ours (w/ ROS)  & 90.85 / 89.47 \\
            (f) Ours (w/ SN) &  92.65 / 92.36 \\
            \toprule[0.8pt]
        \end{tabular}
    \end{small}
    \caption{Effectiveness analysis of Random Object Scaling (AP IoU threshold at 0.5).}
    \vspace{-0.4cm}
    \label{tab:sup_random_object_size}
\end{table}

\begin{table}
    \centering
    \begin{small}
        \begin{tabular}{l|c}
            \bottomrule[1pt]
            Method  & AP$_{\text{BEV}}$ / AP$_{\text{3D}}$ \\
            \hline
            SN  (baseline) & 87.18 / 85.91 \\
            \hline
            ST (w/ SN) &  86.17 / 85.86  \\
            ST (w/ SN) + Triplet        & 86.61 / 85.90 \\
            ST (w/ SN) + Triplet + QAC  & 91.76 / 90.79  \\
            ST (w/ SN) + Triplet + QAC + MEV-C & \textbf{93.57} / \textbf{92.95} \\
            ST (w/ SN) + Triplet + QAC + MEV-C + CDA &  92.65 / 92.36  \\
            \toprule[0.8pt]
        \end{tabular}
    \end{small}
    \caption{Component ablation studies (AP IoU threshold at 0.5). \textbf{ST} represents naive self-training. \textbf{Triplet} means the triplet box partition. \textbf{QAC} indicates the quality-aware criterion. \textbf{MEV-C} is consistency memory ensemble-and-voting. \textbf{CDA} means curriculum data augmentation.}
    \vspace{-0.4cm}
    \label{tab:sup_component_analysis}
\end{table}

\begin{table}[htbp]
    \definecolor{Gray}{gray}{0.9}
    \newcolumntype{a}{>{\columncolor{Gray}}c}
    \centering
    \begin{small}
        \begin{tabular}{lac|acc}
            \bottomrule[1pt]
            $T_{\text{neg}}$ & $T_{\text{pos}}$ & AP$_{\text{BEV}}$ / AP$_{\text{3D}}$ & $T_{\text{neg}}$ & $T_{\text{pos}}$ & AP$_{\text{BEV}}$ / AP$_{\text{3D}}$ \\
            \hline
            0.20 & 0.60  & \textbf{93.34} / \textbf{93.01} & 0.25 & 0.25 & 91.48 / 90.93 \\
            0.25 & 0.60 & 92.65 / 92.36 & 0.25 & 0.30 & 91.17 / 90.70 \\
            0.30 & 0.60 & 93.16 / 92.00 & 0.25 & 0.40 & 92.05 / 91.63 \\
            0.40 & 0.60 & 92.97 / 90.96 & 0.25 & 0.50 & \textbf{92.81} / 92.35 \\
            0.50 & 0.60 & 92.19 / 91.47 & 0.25 & 0.60 & 92.65 / \textbf{92.36} \\
            0.60 & 0.60 & 92.16 / 90.40 & 0.25 & 0.70 & 83.08 / 82.90 \\
            \toprule[0.8pt]
        \end{tabular}
    \end{small}
    \caption{Sensitivity analysis for [$T_{\text{neg}}$, $T_{\text{pos}}$] of triplet box partition (AP IoU threshold at 0.5).}
    \vspace{-0.5cm}
    \label{tab:sup_triplet}
\end{table}

\begin{figure*}
	\centering
	\includegraphics[width=\linewidth]{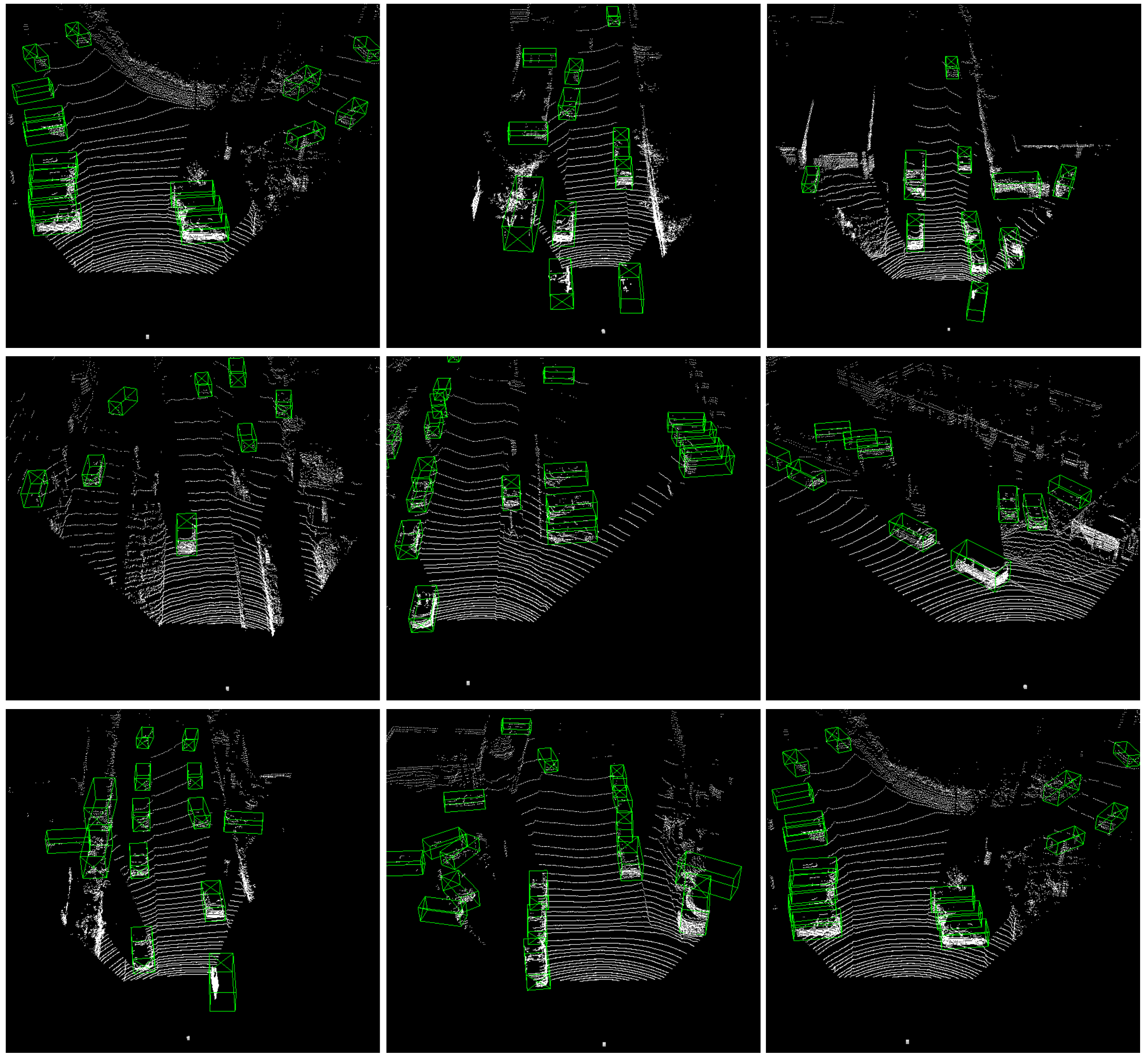}
	\caption{Qualitative results of Waymo $\rightarrow$ KITTI adaptation task.}
	\label{fig:sup_vis_qual}
\end{figure*}
\paragraph{Qualitative Results of ST3D.}
We provide some qualitative results of our proposed ST3D equipped with SN on the KITTI validation set as shown in Fig.~\ref{fig:sup_vis_qual}. Our ST3D (w/ SN) could also predict high-quality object bounding boxes on various scenes with only adaptation and self-training manner.

\begin{table}[htbp]
	\centering
	\vspace{-0.3cm}
	\scalebox{0.95}{
		\begin{small}
			\begin{tabular}{c|c|c|c}
				\bottomrule[1pt]
				Method  & Memory Voting & Merge & AP$_{\text{BEV}}$ / AP$_{\text{3D}}$ \\
				\hline
				ST3D (w/ ME-N) & $\surd$ & Max & \textbf{92.72} / \textbf{92.40} \\
				ST3D (w/ ME-B) & $\surd$ & Max & 92.65 / 92.03 \\
				\hline
				\multirow{4}{*}{ST3D (w/ ME-C)} & $\surd$ & Max & 92.65 / 92.36  \\
				& $\surd$ & Avg & 91.48 / 90.57 \\
				\cline{2-4}
				& $\times$  & Max & 92.66 / 92.22 \\
				& $\times$  & Avg & 90.80 / 90.50 \\
				\toprule[0.8pt]
			\end{tabular}
		\end{small}
	}
    \caption{Ablation studies of memory ensemble (different variants and merge strategies for matched boxes) and memory voting (AP IoU threshold at 0.5). We denote three memory ensemble variants: consistency, NMS and bipartite as ME-C, ME-N, ME-B separately.}
    \vspace{-0.3cm}
	\label{tab:sup_abl_memory_ensemble}
\end{table}

\begin{table}
    \centering
    \vspace{-0.3cm}
    \scalebox{0.92}{
        \begin{tabular}{c|cc|c|c}
            \bottomrule[1pt]
            Method  & World & Object & Intensity & AP$_{\text{BEV}}$ / AP$_{\text{3D}}$ \\
            \hline
            \multirow{6}{*}{ST3D} &$\times$  &$\times$  & - & 83.31 / 66.73 \\
             & $\surd$ & $\times$ & Normal & \textbf{93.62} / \textbf{93.21} \\
             &$\times$  & $\surd$ & Normal & 91.36 / 89.85 \\
             & $\surd$ & $\surd$ & Normal & 93.57 / 92.95 \\
            \cline{2-5} 
            & $\surd$ & $\surd$ & Strong & 92.42 / 91.49 \\
            & $\surd$ & $\surd$ & Curriculum & 92.65 / 92.36 \\
            \toprule[0.8pt]
        \end{tabular}
    }
    \caption{Analysis of data augmentation types and intensities (AP IoU threshold at 0.5).}
    \vspace{-0.3cm}
    \label{tab:sup_abl_aug}
\end{table}

\begin{table*}[htbp]
    \centering
    \vspace{-0.3cm}
    \begin{tabular}{c|c|c|c}
        \bottomrule[1pt]
        Task & Method  & SECOND-IoU & PVRCNN \\
        \hline
        \multirow{5}{*}{Waymo $\rightarrow$ KITTI} & Source Only  & 91.52 / 89.94 & 88.33 / 87.17 \\
        & SN \cite{wang2020train} & 87.18 / 85.91 & 86.32 / 85.72  \\
        \cline{2-4}
        & Ours  & 90.85 / 89.47 & \textbf{92.40 / 92.18} \\
        &Ours (w/ SN) &  \textbf{92.65} / \textbf{92.36} & 91.49 / 90.77 \\
        \cline{2-4}
        & Oracle &  94.08 / 92.28 & 94.97 / 94.85  \\
        \toprule[1pt]
        \bottomrule[1pt]
        \multirow{5}{*}{Waymo $\rightarrow$ Lyft} & Source Only  & 81.82 / 79.73 & 82.38 / 80.45  \\
        & SN \cite{wang2020train} & 81.55 / 78.13& 80.12 / 78.09 \\
        \cline{2-4}
        & Ours  & \textbf{84.44} / \textbf{84.04} & \textbf{84.52} / \textbf{82.61}\\
        & Ours (w/ SN) & 83.98 / 83.40 & 82.21 / 81.70 \\
        \cline{2-4}
        & Oracle &  94.62 / 92.32 &  92.38 / 91.87 \\
        \toprule[1pt]
        \bottomrule[1pt]
        \multirow{5}{*}{Waymo $\rightarrow$ nuScenes} & Source Only  & 43.32 / 37.58  & 40.48 / 36.95 \\
        & SN \cite{wang2020train} & 43.19 / 37.74  & 40.27 / 36.59 \\
        \cline{2-4}
        & Ours & \textbf{43.03} / 38.99 & 40.90 / 38.67 \\
        & Ours (w/ SN) &  42.89 / \textbf{40.21}  & \textbf{41.42} / \textbf{38.99}  \\
        \cline{2-4}
        & Oracle & 63.17 / 58.91 & 61.52 / 58.04  \\
        \toprule[1pt]
        \bottomrule[1pt]
        \multirow{5}{*}{nuScenes $\rightarrow$ KITTI} & Source Only  & 84.32 / 79.18 &  80.88 / 78.47 \\
        & SN \cite{wang2020train} & 48.32 / 46.74 & 66.22 / 65.82 \\
        \cline{2-4}
        & Ours  & 85.59 / 83.62 & 83.75 / 83.64 \\
        & Ours (w/ SN) &  \textbf{86.85} / \textbf{85.65} & \textbf{90.47} / \textbf{90.25} \\
        \cline{2-4}
        & Oracle &  94.08 / 92.28 & 94.97 / 94.85  \\
        \toprule[0.8pt]
    \end{tabular}
    \caption{Result of different adaptation tasks. We report AP of the car category in $\text{AP}_{\text{BEV}}$ and $\text{AP}_{\text{3D}}$ at \textbf{IoU = 0.5}. The reported result is for the moderate case on the adaptation tasks with KITTI as target domain, and is the overall result for other adaptation tasks.}
    \label{tab:sup_SOTAcomparison}
\end{table*}

\begin{table*}[htbp]
    \centering
    
    \begin{small}
    \scalebox{1}{
        \begin{tabular}{l|c|c|c|c|c}
            \bottomrule[1pt]
            AP$_{\text{BEV}}$ / AP$_{\text{3D}}$  & nuScenes$\rightarrow$ Waymo & nuScenes $\rightarrow$ Lyft & Lyft $\rightarrow$ KITTI & Lyft $\rightarrow$ Waymo & Lyft $\rightarrow$ nuScenes\\
            \hline
            Source Only & 20.47 / 09.39 & 39.79 / 17.29 & 77.55 / 55.39 & 51.87 / 37.89 & 30.43 / 17.52 \\
            SN & 19.83 / 03.17 & 34.65 / 14.15 & 81.08 / 65.01 & 51.85 / 39.42 & 30.18 / 18.13 \\
            \hline
            ST3D & \textbf{49.29} / \textbf{23.86} & \textbf{58.12} / \textbf{33.48} & 85.03 / 68.92 & 56.64 / 40.89 & \textbf{33.26} / 19.76 \\
            ST3D (w/ SN) & 25.24 / 11.00 & 51.20 / 26.41 & \textbf{85.10} / \textbf{71.42} & \textbf{57.76} / \textbf{42.89} & 32.89 / \textbf{21.49} \\
            \hline
            Oracle & 65.01 / 51.12 & 84.47 / 68.78 & 83.29 / 73.45 & 65.01 / 51.12 & 51.88 / 34.87 \\
            \toprule[0.8pt]
        \end{tabular}
    }
    \end{small}
    \caption{ Result of other five adaptation tasks. Notice that we sample $\frac{1}{20}$ of Waymo training frames and $\frac{1}{10}$ of Waymo validation frames when Waymo serves as target domain.}
    \label{tab:sup_extra_exp}
    \vspace{-0.5cm}
\end{table*}

\section{Experimental Result on More Tasks.}
\label{sec:extra_tasks}
Our experiments in the main paper are designed to cover most practical scenarios (across different LiDAR beam ways and from label-rich domains to label insufficient domains), and we also rule out some ill-posed settings, such as we do not consider KITTI and Lyft as source domain since KITTI lacks of ring view annotations and Lyft has very difference annotations in our main paper (see Sec.~\textcolor{red}{S2.1} in supplementary materials).
However, to validate the effectiveness of our method, we further conduct 5 extra experiments. Tab.~\ref{tab:sup_extra_exp} shows that, without tuning hyper-parameters, ST3D still achieves promising results on these five adaptation tasks.

\end{document}